\documentclass{article}

\PassOptionsToPackage{round, compress}{natbib}
\usepackage[preprint]{neurips_2026}

\usepackage[utf8]{inputenc} % allow utf-8 input
\usepackage[T1]{fontenc}    % use 8-bit T1 fonts
\usepackage{url}            % simple URL typesetting
\usepackage{booktabs}       % professional-quality tables
\usepackage{amsfonts}       % blackboard math symbols
\usepackage{nicefrac}       % compact symbols for 1/2, etc.
\usepackage{microtype}      % microtypography
\usepackage{xcolor}         % colors
\definecolor{darkblue}{rgb}{0, 0, 0.2}
\definecolor{darkgreen}{rgb}{0, 0.2, 0}
\definecolor{myteal}{HTML}{1F4F66}
\usepackage[colorlinks=true,
            linkcolor=darkblue,
            citecolor=darkgreen,
            urlcolor=darkblue,
            backref=page]{hyperref}       % hyperlinks
\usepackage{amsmath,amsfonts,bm}

\def\eqref#1{equation~\ref{#1}}
\def\1{\bm{1}}

\def\vk{{\bm{k}}}

\def\vo{{\bm{o}}}

\def\vq{{\bm{q}}}

\def\vv{{\bm{v}}}

\def\mG{{\bm{G}}}

\def\mK{{\bm{K}}}

\def\mV{{\bm{V}}}

\DeclareMathAlphabet{\mathsfit}{\encodingdefault}{\sfdefault}{m}{sl}
\SetMathAlphabet{\mathsfit}{bold}{\encodingdefault}{\sfdefault}{bx}{n}

\def\gC{{\mathcal{C}}}

\def\gN{{\mathcal{N}}}

\def\gR{{\mathcal{R}}}

\newcommand{\R}{\mathbb{R}}

\usepackage{soul}           % underline \ul
\usepackage{tabularx}
\usepackage{graphicx}    % \resizebox
\usepackage{colortbl}
\usepackage{xcolor}
\usepackage{array}
\usepackage{mathtools}
\usepackage{multirow}
\usepackage{xurl}
\usepackage{titletoc}
\usepackage{amsthm,thmtools}

\definecolor{ourrow}{HTML}{E8F2F9} 
\definecolor{seerrow}{HTML}{FFFBE6}

\usepackage[suppress]{color-edits}
\addauthor[Ting-Yun]{tc}{magenta}
\addauthor[Harvey]{hf}{blue}
\addauthor[Deqing]{df}{red}
\addauthor[Chenghao]{ch}{purple}
\usepackage{cleveref}
\newcommand{\ourmethod}{\textsc{VaSE}}
\newcommand{\ourmethodattn}{\ourmethod{}-\textsc{AttnV}}
\newcommand{\ourmethoddkv}{\ourmethod{}-\textsc{DKV}}
\title{Value-Aware Stochastic KV Cache Eviction for Reasoning Models}

\newcommand{\uchicago}{\ensuremath{\textcolor[HTML]{800000}{\boldsymbol\chi}}}
\newcommand{\usc}{\ensuremath{\textcolor[HTML]{990000}{\boldsymbol\sigma}}}
\newcommand{\aspace}{\hspace{4mm}}
\author{%
\textbf{Ting-Yun Chang}$^{\star\usc}$\aspace
\textbf{Harvey Yiyun Fu}$^{\star\uchicago}$\aspace
\textbf{Deqing Fu}$^{\star\usc}$\aspace
\textbf{Chenghao Yang}$^{\uchicago}$\\
\textbf{Jesse Thomason}$^{\usc}$\aspace
\textbf{Robin Jia}$^{\usc}$\\
$^{\usc}$University of Southern California\aspace
$^{\uchicago}$University of Chicago\\
\texttt{\{tingyun,deqingfu,jessetho,robinjia\}@usc.edu}\\\texttt{\{harveyfu,chenghao\}@uchicago.edu}
}
\makeatletter
\renewcommand{\@noticestring}{$^\star$ Equal contribution.}
\makeatother

\usepackage{titlesec}
\titlespacing\paragraph{0pt}{2pt}{4pt plus 2pt minus 2pt} 
\titlespacing\section{4pt}{4pt}{0pt}
\begin{document}

\maketitle

\begin{abstract}
  %We introduce \textbf{\ul{V}}alue-\textbf{\ul{a}}ware \textbf{\ul{S}}tochastic KV Cache \textbf{\ul{E}}viction (\ourmethod).
 Reasoning models improve accuracy through extended chains of thought, but their long outputs create a memory and compute bottleneck. 
 KV cache eviction methods reduce this cost by evicting unimportant key-value pairs from the cache, yet they often yield worse accuracy than selection-based sparse attention alternatives, which keep the full KV cache. 
 We identify key factors crucial to KV cache eviction accuracy. First, a small fraction of value states have abnormally large magnitudes, and evicting them causes catastrophic failure where models enter repetitive reasoning loops. Second, introducing stochasticity during eviction improves accuracy by increasing cache diversity. Based on these findings, we propose \textbf{V}alue-\textbf{a}ware \textbf{S}tochastic KV Cache \textbf{E}viction (\ourmethod), a training-free recipe that protects large-magnitude value states and promotes diverse eviction decisions. Across six reasoning tasks, Qwen3 models using \ourmethod\ with $4\times$ KV cache compression yield higher average accuracies than SOTA selection method at the same sparsity, while outperforming the strongest eviction method by more than 4\%.
 Overall, \ourmethod\ bridges the gap between efficiency and accuracy, supporting FlashAttention2 and enabling a static memory footprint for reasoning models.
 %enabling a static memory footprint that selection methods cannot offer. 
 %\ourmethod{} generalizes as a recipe, with its key factors improving SOTA eviction methods by more than 4\% on average.
 %Overall, our work provides a simple yet effective recipe for accurate and memory-efficient KV cache eviction in long-context reasoning.
\end{abstract}

\begin{figure}[t]
    \centering
    \includegraphics[width=\linewidth]{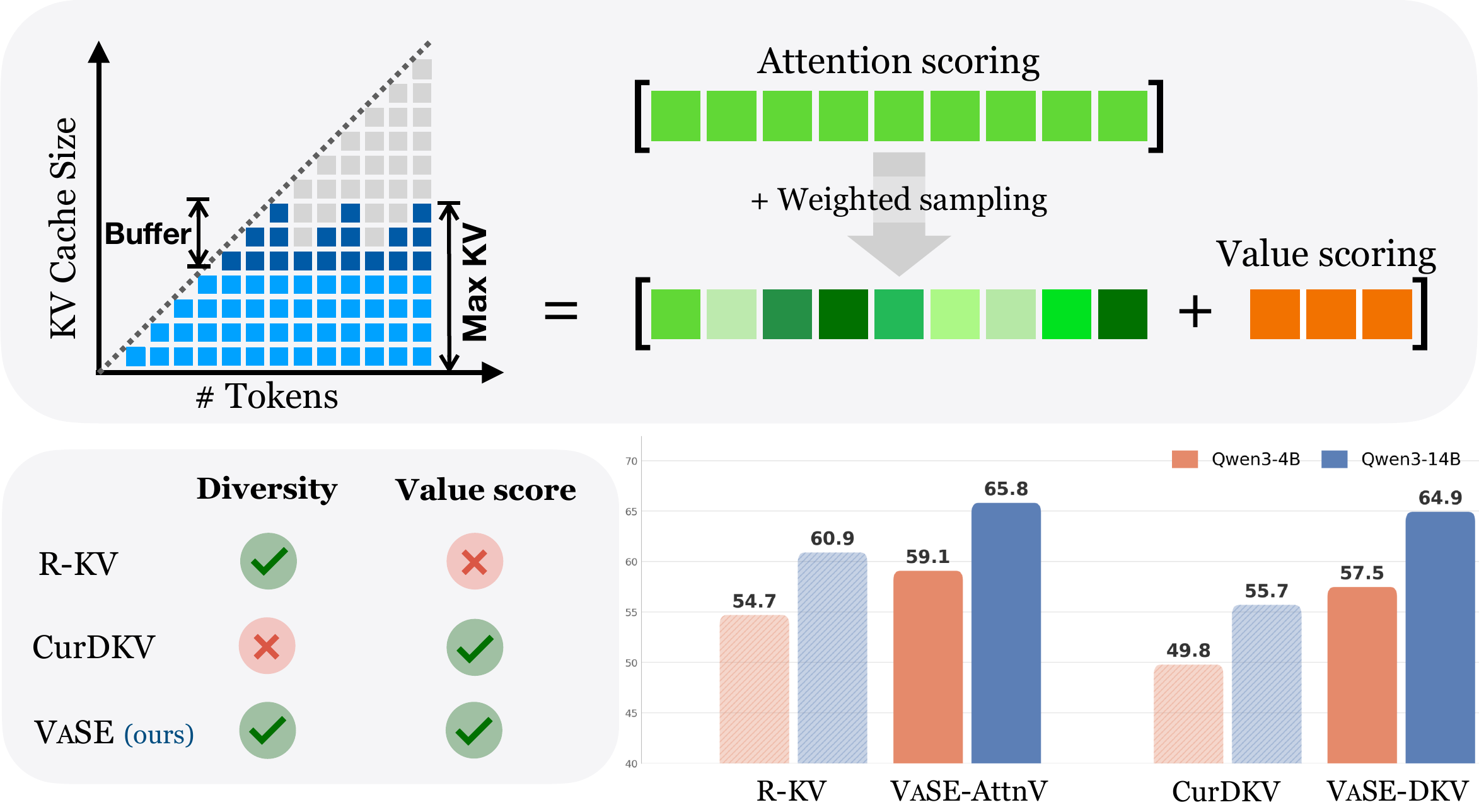}
    \caption{\textit{Top}: \ourmethod{} is a KV Cache eviction method that combines stochastic sampling with value-state magnitude scoring to retain diverse and important KV pairs under a fixed KV cache budget. 
    \textit{Bottom}: By integrating both stochasticity and value awareness, \ourmethod{} outperforms baseline methods that use either signal alone, improving average pass@1 accuracy across various reasoning tasks.}
    \label{fig:main_figure}
\end{figure}
\newcommand{\KVBudget}{K}
\newcommand{\KVBuffer}{B}
\newcommand{\ValScore}{f}
\newcommand{\KVFull}{Full}
\newcommand{\KVAttn}{SnapKV}
\newcommand{\KVRkv}{R-KV}
\newcommand{\KVCur}{CurDKV}
\newcommand{\KVAttnV}{SnapKV+V}
\newcommand{\KVSeerR}{SeerAttention-R}
\newcommand{\Vslots}{N_{v}}
\newcommand{\Fullavg}{N_{\text{avg}}}
\newcommand{\Rule}{pruning rule}
\newcommand{\Ssnap}{\bar\alpha_i}
\newcommand{\Range}{\mathrm{Range}}
\newcommand{\Scale}{s_\vv}

\section{Introduction} \label{sec:intro}
Reasoning models \citep{openai2024o1,qwen2024qwq,guo2025deepseek} leverage test-time compute \citep{snell2025scaling} to improve accuracy by generating extended chains of thought before producing a final answer.
While this approach yields high accuracy on complex tasks, it introduces an efficiency bottleneck at test time because the reasoning models tend to generate unnecessary long outputs. 
For example, \citet{chen2024not} show that models may overthink a simple arithmetic question, using more than 900 tokens to answer ``2 + 3 = ?’’.
As a consequence, an auto-regressive large language model that stores the key-value representations of every past token incurs substantial memory and computational overhead as the sequence length grows.

Sparse attention \citep{zhao2019explicit,beltagy2020longformer,zaheer2020big,kitaevreformer} addresses this bottleneck by attending to only a subset of previous tokens; such approaches fall into two broad categories.
KV cache \emph{selection} methods \citep{ribarsparq,tang2024quest,yang2025tidaldecode,gao2026sparse} keep the full KV cache in memory but activate only a sparse subset of KV pairs at each decode step. These methods reduce compute and memory movement, but their memory footprints scale linearly with sequence length.
In contrast, KV cache \emph{eviction} methods \citep{zhang2023h2o,gemodel,devoto-etal-2024-simple,oren-etal-2024-transformers} permanently discard low-importance KV pairs once the cache reaches a predefined budget.
%, reducing the memory from $O(T)$ to $O(\KVBudget)$ and the compute from $O(T^2)$ to $O(T \times \KVBudget)$, where $T$ is the number of tokens.
While eviction methods \citep{li2024snapkv,cai2025rkv} suffer larger accuracy degradation on reasoning tasks compared to their selection-based counterparts, they can yield a static memory footprint and better throughput.

In this paper, we reveal two key findings to improve KV cache eviction methods on reasoning models.
First, the magnitudes of value states show strongly skewed distributions, with a small fraction of tokens having abnormally large vector magnitudes (see Figure \ref{fig:violin_evict}).
Those large-magnitude value states are crucial for maintaining task accuracies: evicting them causes the models to enter repetitive loops, where they endlessly re-examine the context or generate nonsensical outputs without reaching a final answer (see Figure \ref{fig:loop} for the examples).
Second, we find that introducing stochasticity during KV cache eviction effectively improves reasoning task accuracies. We attribute this to improved diversity of the retained KV pairs, which together yield a more representative coverage of the full context. Our case study on GSM8K demonstrates that incorporating value scoring boosts Qwen3-4B's accuracy by as much as 16.2\%, while introducing stochasticity yields an additional 4.7\% improvement.

Building on these findings, we propose \textbf{V}alue-\textbf{a}ware \textbf{S}tochastic KV Cache \textbf{E}viction (\ourmethod{}), a training-free KV cache eviction recipe that prioritizes large-magnitude value states and introduces stochasticity during eviction.
We target the decoding step of reasoning models, by applying \ourmethod{} to Qwen3 models \citep{yang2025qwen3} and evaluating on six tasks spanning math, code generation, and science question answering.
On average, \ourmethod{} outperforms the previously SOTA eviction method R-KV \citep{cai2025rkv} by 4.4\% on Qwen3-4B and 4.9\% on Qwen3-14B.
When combined with CurDKV \citep{sengupta2025cur}, stochastic sampling further boosts its average accuracy by 7.7\% and 9.2\% on the two models, respectively.
\ourmethod{} even slightly surpasses the average accuracy of SeerAttention-R \citep{gao2026sparse}, a strong selection method whose KV cache memory grows linearly with the sequence length, on both models while using static memory.
In addition, we benchmark the actual throughput and peak memory of different methods, showing that \ourmethod{} achieves higher throughput and a lower memory footprint than the R-KV eviction baseline across token budgets. Our results match closely with the theoretical memory compression ratio.

%We further connect our findings to KV cache quantization by showing that large-magnitude value states also lead to disproportionately large reconstruction errors under per-token quantization \citep{liu2024kivi}.
Furthermore, our insights generalize to other efficiency methods like KV cache quantization; we show that large-magnitude value states also lead to disproportionately large reconstruction errors under per-token quantization \citep{liu2024kivi}.
Together, we demonstrate that value-state magnitude and stochasticity are fundamental axes of KV cache management, with our \ourmethod\ recipe bridging the efficiency-accuracy gap left by prior sparse attention methods.\footnote{Our code is available at \url{https://github.com/terarachang/VaSE}.}

\section{Problem Setup} \label{sec:problem}

%Two families of methods address the per-step cost of attending to a long KV cache during decoding. KV cache \textit{selection} keeps the full cache in memory and activates only a sparse subset of pairs at each decoding step. 
We formalize the decode-phase KV cache eviction setting by drawing the distinction between selection-based and eviction-based methods, and introduce several representative eviction baselines.

\paragraph{Selection-based method.}
Selection-based sparse attention methods retain the full KV cache but activate only a sparse subset of KV pairs at each decoding step. A representative example is SeerAttention-R \citep{gao2026sparse}, which uses a trained attention gate to predict attention sparsity patterns and selectively activates different KV cache blocks during inference, thereby reducing computational complexity and memory movement. Compared to eviction-based methods, SeerAttention-R retains all KV pairs and the memory footprint scales linearly as $\mathcal{O}(T)$ with sequence length $T$. 
% SeerAttention-R \citep{gao2026sparse} is a selection-based sparse attention method that trains a gating mechanism to predict attention sparsity patterns.
% During inference, the gate selectively activates different KV cache blocks at each decoding step, thereby reducing both computational complexity and the memory bandwidth bottleneck of full attention.
% Comparing to eviction-based methods, SeerAttention-R does not reduce the memory footprint of KV cache, as all pairs are retained. 
% Unlike selection-based methods such as SeerAttention-R \citep{gao2026sparse}, which retain the full KV cache and sparsify attention at each step, eviction methods enables a constant memory footprint by permanently removing KV pairs from the cache. Each eviction decision is irreversible and can affect all subsequent decoding steps.
% Sparse attention methods fall into two broad categories: selection-based and eviction-based.

% \paragraph{Selection-based method.}
% %\dfcomment{talk about sparsity here.}

\paragraph{Eviction-based method.}
KV cache eviction methods permanently discard low-importance key-value pairs once the cache size reaches a predefined budget, reducing memory cost at the risk of irrecoverable information loss. 
We specifically focus on eviction methods for the decoding step of autoregressive reasoning models, as these models can generate over $10{,}000$ tokens for a math question that has fewer than 200 prompt tokens (see \Cref{tab:dataset_tokens} for statistics).

Formally, let $N$ be the total KV cache budget, and $d$ be the dimension of keys and values.
As the KV cache budget fills up at the token index $t$, we have $N$ KV pairs $(\vk_i, \vv_i)$ for $i = 1, \ldots, N$, where $\vk_i \in \R^d, \vv_i \in \R^d$.
Given the current query $\vq_t \in \R^d$, the attention head computes its output as a weighted sum over the $N$ cached values,
\begin{equation}
  \vo_t \;=\; \sum_{i \leq N} \alpha_i^{(t)} \, \vv_i,
  \qquad
  \alpha_i^{(t)} \;=\;
      \frac{\exp\!\left(\vq_t^\top \vk_i / \sqrt{d}\right)}
           {\sum_{j \leq N} \exp\!\left(\vq_t^\top \vk_j / \sqrt{d}\right)},
  \label{eq:attn}
\end{equation}
where $\alpha_i^{(t)}$ is the attention weight assigned to token $i$,
obtained by applying softmax to the scaled dot products between the current
query and all cached keys \citep{vaswani2017attention}.

\paragraph{Periodic eviction with budget $K$ and buffer $B$.}
We adopt the periodic-eviction framework proposed by \citet{cai2025rkv} and \citet{song2025reasoning}, which targets decode-phase compression.
The framework consists of a persistent budget of $K$ tokens and a buffer of size $B \ll K$ holding the most recent tokens, capping the total cache size at $N = K + B$ tokens.
Since each decoding step adds one KV pair to the cache, the buffer fills every $B$ steps and triggers an \emph{eviction step}: the eviction operator chooses $B$ pairs to discard, clears the buffer, and restores the cache to size $K$ (See the upper-left panel in Figure~\ref{fig:main_figure}). 
Concretely, the most recent $B$ KV pairs in the buffer are protected from eviction. 
A \emph{scoring function} then assigns a scalar importance score to each of the remaining $K$ candidates, and the $B$ candidates with the lowest importance scores will be discarded.
%\hfcomment{maybe give an example here} assigns a scalar importance score to each pair, and then a \emph{\Rule} \hfcomment{maybe give an example here} determines an evicted set based on the importance scores.
Below, we describe the scoring functions of three prior eviction methods, SnapKV \citep{li2024snapkv}, R-KV  \citep{cai2025rkv},
and \KVCur \citep{sengupta2025cur}.
Following common practice, KV pairs are scored independently for each attention head and layer.
To simplify notation, we omit layer and head indices hereafter.

\paragraph{SnapKV.}
\citet{li2024snapkv} rank candidate pairs by their average attention score over a window of recent queries, where we set the window size to be the same as the buffer size $B$.\footnote{Following \citet{li2024snapkv}, we apply average pooling to $\Ssnap$ to smooth the attention scores. Following the decode-phase SnapKV variant introduced by \citet{cai2025rkv}, we cache the most recent queries and compute non-causal attention scores.}
%the key matrix $\mK \in \R^{N \times d}$ and value matrix $\mV \in \R^{N \times d}$, with rows $\vk_i := \mK_{i,:}$ and $\vv_i := \mV_{i,:}$ for $i = 1, \ldots, N$.
%SnapKV \citep{li2024snapkv} ranks candidate pairs by their average attention score over a window of recent queries.
%and score each candidate by the average
%at eviction step $t$ by averaging its attention probability over the $B$ buffered queries:
\begin{equation}%^{(t)}
    \Ssnap \coloneqq \frac{1}{B} \sum_{\tau = t - B + 1}^{t} \alpha_i^{(\tau)},
    \label{eq:snapkv-score}
\end{equation}
where $i$ is the KV pair index, $t$ is the eviction timestep, and $\alpha_i^{(\tau)}$ is the attention score defined in Eq.~\ref{eq:attn}.
SnapKV uses keys for the attention computation, but does not consider values in the scoring function.

\paragraph{R-KV.} \citet{cai2025rkv} specifically target redundant tokens in reasoning models.
They augment the SnapKV score $\Ssnap$ with a
redundancy penalty $r_i$ to promote diversity among retained KV pairs:
\begin{equation}
s_i^{\text{RKV}} \coloneqq \lambda \cdot \Ssnap - (1 - \lambda) \cdot r_i,
\qquad \text{where} \quad
r_i = \operatorname{Softmax}\!\left(
  \frac{1}{K} \sum_{j \neq i} \operatorname{CosSim}(\vk_i, \vk_j)
\right),
\end{equation}
where $\lambda \in [0,1]$ is a hyperparameter. A high $r_i$ indicates that
token $i$ has a high cosine similarity to many cached tokens; subtracting it discourages
choosing redundant pairs.

\paragraph{\KVCur.}
\citet{sengupta2025cur} observe the limitations of scoring KV pairs with attention alone, which ignores value states.
They compute leverage scores for keys and values respectively via approximate CUR matrix decomposition and combine them into a single importance score.
Originally designed for the prefill stage, we extend \KVCur\ for the decode phase with the periodic eviction framework.
Specifically, \KVCur\ draws a Gaussian projection $\mG \in \R^{d \times r}$ at the start of generation, with each entry $G_{ab} \overset{\textnormal{iid}}{\sim} \gN(0, 1/r)$, and rank candidates by the product of key and value scores:
\begin{equation}\label{eq:proj-scores}
    s_i^{\text{CUR}} \coloneqq \ell_i^{(K)}(\mG) \cdot \ell_i^{(V)}(\mG), \qquad \text{where} \quad \ell_i^{(K)}(\mG) \;:=\; \|\mG^\top \vk_i\|_2^2, \quad \ell_i^{(V)}(\mG) \;:=\; \|\mG^\top \vv_i\|_2^2.
\end{equation}
%The \Rule\ evicts candidates with the lowest $s_i^{\text{CUR}}$ score.
\KVCur\ considers value states during scoring, but it does not promote diversity among KV pairs. 

\section{Methodology} \label{sec:method}
\begin{figure}[t]
    \centering
    \includegraphics[width=\linewidth]{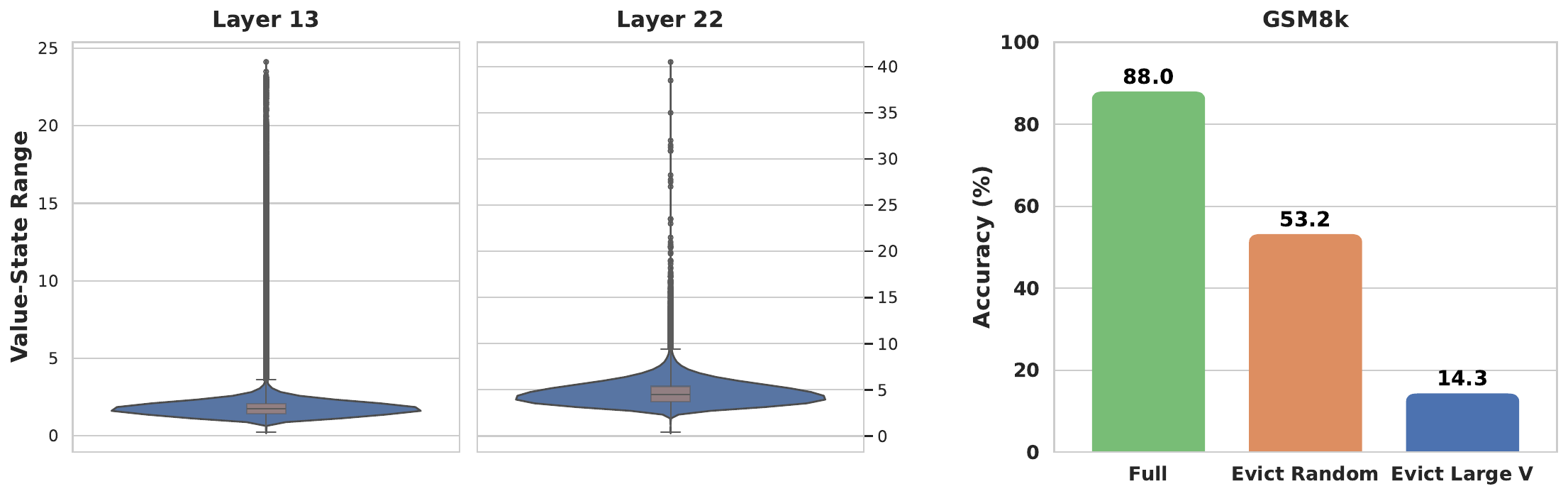}
    \caption{\textit{Left}: $\Range$ distribution of the value states. The violin plots show the presence of extreme magnitude outliers at different layers. \textit{Right}: Evicting the large magnitude outliers causes accuracy to collapse to $14.3\%$, greatly underperforming a random eviction baseline at the same token budget, suggesting that these large-magnitude value states are crucial to model accuracy.}
    \label{fig:violin_evict}
\end{figure}
We hypothesize that incorporating value states and diversity into KV cache scoring is essential for preserving model accuracy under eviction.
Inspired by the scoring function of \KVCur, which favors value vectors with larger magnitudes, and by the finding of \citet{sun2024massive} that massive activations in LLMs are indispensable, we first study the effect of large-magnitude value states on eviction accuracy. 
Based on the analysis, we propose a framework that prioritizes those large-magnitude value states within the scoring function and introduces stochasticity to promote diversity.

\subsection{Significance of Value States with Large Magnitudes}
\label{sec:v_signficance}
We first analyze the distribution of value-state magnitudes, where we extract the full KV cache from Qwen3-4B while generating responses to GSM8K examples \citep{cobbe2021gsm8k}.
We define the magnitude\footnote{We experiment with three variants, $\mathrm{Range}$, $L_2$ norm, and variance, to capture the magnitude and variety of value vectors in Appendix~\ref{sec:app_vmag}. We find that these variants are positively correlated and lead to comparable accuracies.} of a cached value vector $\vv \in \R^d$ as:
\begin{equation} \label{eq:value-score}
 \mathrm{Range}(\vv) \;:=\; \max_{j \in [d]} v_{j} \;-\; \min_{j \in [d]} v_{j}
\end{equation}
where $v_j$ is the $j$-th entry of $\vv$, and  ${\mathrm{Range}}(\vv)$ is chosen over $L_2(\vv)$ for its connection to quantization (see \S\ref{sec:quantization}).
\Cref{fig:violin_evict} shows that the distribution of $\Range(\vv)$ has large outliers over layers (see Appendix~\ref{sec:app_vmag} for more layers).
%We measure the Fisher skewness $g = \frac{m_3}{m_2^{3/2}}$ of the value-state magnitude, where $m_i$ is the $i$-th central moment, and $g > 0$ means a longer right tail. \dfcomment{I'll try to define skewness better here. }.
%We find that every layer has a positive skewness $g$, with some exceeding 5, showing an extreme right tail of outliers. \hfcomment{not sure of skewness is reused in later parts of the results. readers may not get an sense of how large 5 is}
We hypothesize that these large-$\Range$ outliers are critical to the model accuracy.
Because attention output is computed as a weighted summation of values (Eq. ~\ref{eq:attn}), those values with larger magnitudes have a disproportionate influence on the output. 

To validate this hypothesis, we purposefully evict the
$B$ largest KV pairs ranked by $\Range(\vv_i)$. 
This reduces GSM8K accuracy to just
14.3\%, under-performing random eviction by 
38.9\% at the same 512-token budget (\Cref{fig:violin_evict}; \textit{Right}).
Further inspection of the model outputs reveals that the model tends to enter endless loops of self-reflection and never reaches a conclusion; it also tends to generate non-grammatical sentences (see~\Cref{fig:loop}).
Prior work \citep{guo-etal-2024-attention,guoactive} observes that low-magnitude values co-occur with attention sinks \citep{xiao2024efficient}, which are crucial to eviction methods. In this paper, we demonstrate that large-magnitude values are equally important and should be up-weighted within the scoring function (\S \ref{sec:ablation}).

\begin{table}[t]
\setlength{\tabcolsep}{8pt}
\renewcommand{\arraystretch}{1.3}
\noindent\resizebox{\linewidth}{!}{%
\begin{tabular}{@{}lllll@{}}
\toprule
\textbf{Method} & \textbf{Family} & \textbf{Key score} & \textbf{Value score} & \textbf{Diversity} \\
\midrule
SeerAttention-R \citep{gao2026sparse} & Select & Attn Gate                            & --
%\dfcomment{NA looks ugly here.}
& -- \\
\midrule
SnapKV \citep{li2024snapkv}  & Evict  & $\bar\alpha_i$                                  & --                         & -- \\
\KVRkv\ \citep{cai2025rkv}  & Evict  & $\lambda \cdot \bar\alpha_i - (1 - \lambda) \cdot r_i$ & --                        & redundancy $r_i$ \\
\KVCur\ \citep{sengupta2025cur}   & Evict  & $\ell_i^{(\mK)} = \|(\mK\mG)_i\|_2^2$  & $\ell_i^{(\mV)} = \|(\mV\mG)_i\|_2^2$ & -- \\
\midrule
\rowcolor{ourrow}
\textbf{\ourmethodattn (ours)} & Evict & $\bar\alpha_i$ & $\mathrm{Range}(\vv_i)$ & sample $\bar\alpha_i$ \\
\rowcolor{ourrow}
\textbf{\ourmethoddkv (ours)}  & Evict & $\ell_{i, t}^{(\mK)} = \|(\mK\mG_t)_i\|_2^2$ & $\ell_{i, t}^{(\mV)} = \|(\mV\mG_t)_i\|_2^2$ & sample $\mG_t$ \\
\bottomrule
\end{tabular}%
}
\vspace{0.2mm}
\caption{Design space of sparse attention methods. Among eviction methods, prior work covers at most two of the three axes (key-based scoring, value-based scoring, diversity); \ourmethod{} is the first to combine all three, instantiated as \ourmethod{}-AttnV and \ourmethod{}-DKV.}
\label{tab:design-space}
\end{table}
\subsection{\ourmethod{}: Value-Aware Stochastic Eviction}
We propose \ourmethod{} (Value-aware Stochastic Eviction), a training-free eviction recipe designed around two principles: upweighting large-magnitude value states in the scoring function, and introducing stochasticity to promote diversity in the retained cache.
We propose two \ourmethod{} variants that apply our recipe to SnapKV and CurDKV, respectively.

\paragraph{\ourmethodattn.} \ourmethodattn\ reserves part of the token budget for large-magnitude value states and applies stochastic sampling on top of SnapKV.
Recall that $K$ is the persistent budget and $B$ is the buffer size (\S~\ref{sec:problem}). 
We first rank each KV pair by $\mathrm{Range}(\vv_i)$ (Eq.~\ref{eq:value-score}), where $i\in[K]$ indexes the non-buffer tokens.
Given a value reservation budget $\Vslots < K$, we select the $\Vslots$ candidates with the largest $\Range(\vv_i)$ and unconditionally retain them as the reserved set $\gR_\mV$.
We denote the set of tokens in the buffer as $\gR_B$, forming the full reserved set $\gR = \gR_\mV \cup \gR_B$.
The remaining $K - |\gR|$ slots are then filled by sampling from the SnapKV attention distribution, with weights proportional to $\bar{\alpha}_i$ (Eq.~\ref{eq:snapkv-score}).
This preserves SnapKV's core idea of favoring high-attention tokens, while replacing its hard topk selection with a soft, probabilistic one.
%Concretely, let $\gC = [K] \setminus \gR_\mV$ be the candidates not held by value reservation. The remaining $K - |\gR|$ slots are drawn from $\gC$ by weighted sampling without replacement, with weights proportional to $\bar{\alpha}_i$, forming the final retained set. 

%At step $t$, let $\pi_i^{(t)}$ denote the probability of token $i$ being retained, conditional on the candidate set $\gC_t$ and the current attention weights $\{\bar\alpha_j^{(t)}\}$. 
Stochastic sampling offers a key advantage over topk in terms of token retention flexibility.
Let $\pi_i^{(t)}$ denote the probability of token $i$ being retained at step $t$.
After $T$ eviction steps, the probability that token $i$ is kept in the KV cache can be factorized as $\prod_{t=1}^{T} \pi_i^{(t)}$. 
%Under Topk, a single $\pi_i^{(t)}=0$ permanently discards any token falling below the cutoff.
%In contrast, sampling without replacement guarantees that for any token with a positive attention weight, $\pi_i^{(t)} \geq 1 - (1 - \Ssnap^{(t)}/Z_t)^{K - |\gR|} > 0,$ where $Z_t = \sum\nolimits_{j \in \gC_t} \bar\alpha_j^{(t)}$ and $\gC_t = [K] \setminus \gR_\mV^{(t)}$ is the candidate set for sampling.
%Thus, every token with a non-zero attention weight remains eligible at a future step.
Under topk, any token below the cutoff is evicted with certainty ($\pi_i^{(t)}=0$). 
In contrast, under sampling, every factor is strictly positive, $\pi_i^{(t)} \geq 1 - (1 - \Ssnap^{(t)}/Z_t)^{K - |\gR|} > 0,$ where $Z_t = \sum\nolimits_{j \in \gC_t} \bar\alpha_j^{(t)}$ and $\gC_t = [K] \setminus \gR_\mV^{(t)}$ is the candidate set for sampling; therefore, each token has a nonzero probability of being retained across all $T$ steps.

% The rate of a token being retained across $T$ steps can be factorized as $\Pr[\tau_i > T] = \prod_{t=1}^{T} \pi_i^{(t)}$.
% Therefore under top-$K$, 
% Under top-$K$, any eviction step where the token falls below the cutoff removes it with probability one, so its survival probability over any longer horizon becomes zero.
% Under sampling, every factor remains strictly positive whenever the token has nonzero sampling weight: below-cutoff tokens therefore retain a positive finite-horizon survival probability, rather than being killed at the first step where their rank drops below $K$.

%Sampling preserves the SnapKV signal that high-attention tokens are more likely to be retained, but never assigns probability one to any single candidate.
% Two near-duplicates with $\Ssnap \approx \bar\alpha_j$ no longer have correlated fates: under deterministic top-$K$ either both win the tie-break and are retained, or both lose and are evicted; under sampling, the two are drawn approximately independently, so cache budget is less likely to be spent on redundant copies.
%A token with $\Ssnap$ just below the deterministic top-$K$ cutoff is evicted with probability one under SnapKV at every eviction step, whereas under sampling it is retained with some small probability. Over the thousands of eviction steps in a long reasoning trace, this lets the cache explore a wider set of mid-attention tokens rather than committing to the same boundary decision at every step.

\paragraph{\textsc{\ourmethod{}-DKV}.} \textsc{\ourmethod{}-DKV} inherits value-state awareness directly from \KVCur's product score (Eq.~\ref{eq:proj-scores}), where the value score $\ell_i^{(V)}(\mG)$ upweights large-magnitude values.
  On top of this, we introduce stochasticity through the Gaussian matrix $\mG$ at each eviction step $t$ to promote diversity among the retained KV pairs.
  Specifically, at step $t$ we resample a random projection $\mG_t \in \R^{d \times r}$ independently of all past projections, and compute the leverage scores
  following Eq.~\ref{eq:proj-scores}.
  This improves upon \KVCur, which uses a fixed $\mG$ across all steps: tokens with certain representations are then consistently assigned low scores and deterministically evicted.
  Resampling $\mG$ avoids this failure mode by applying an independent scoring criterion at each eviction step such that no token would be permanently disadvantaged by a single projection.
% Resampling implements stochasticity at the level of the score rather than the \Rule\: even though top-$K$ is deterministic given the score, the score itself is fresh at every eviction, so a token whose product score sits near the eviction threshold is retained with probability bounded away from zero at every eviction under fresh draws of $\mG_t$, whereas under \KVCur's single $\mG$ a single unfavorable draw keeps the same token below threshold for the entire generation.

\section{Experiments} \label{sec:experiments}

\begin{figure}[t]
    \centering
    \includegraphics[width=\linewidth]{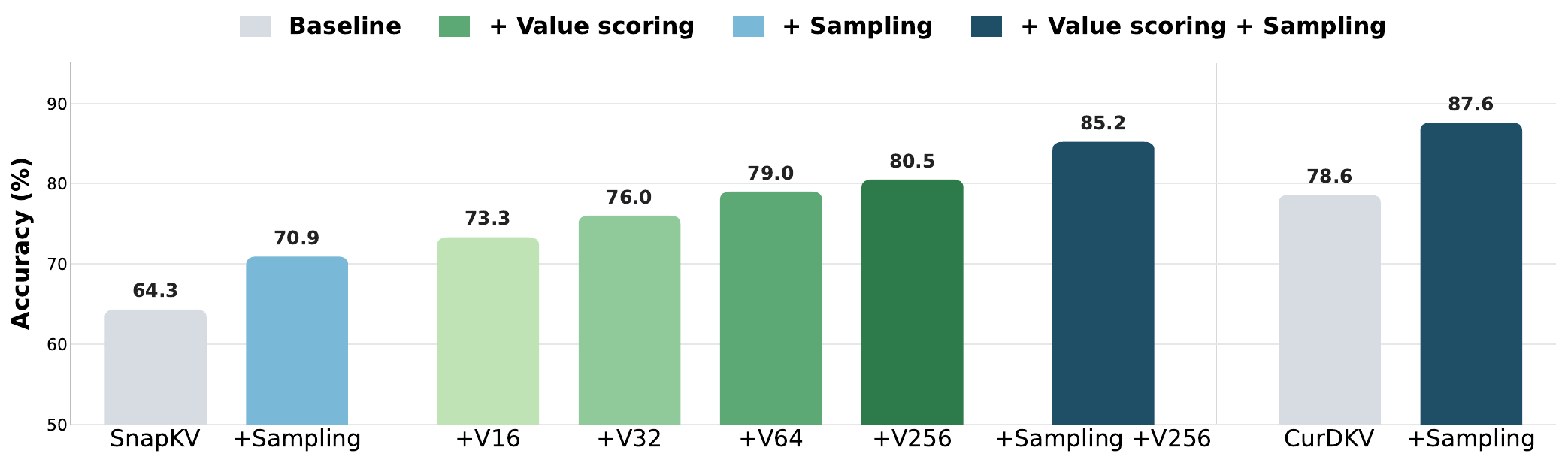}
    \caption{Accuracy of Qwen3-4B on GSM8K with a KV cache budget of $\KVBudget{=}512$ ($\sim$$4\times$ compression). \textbf{+V16} indicates reserving 16 slots in the budget $K$ for value scoring. By combining stochastic sampling with value scoring, our methods \colorbox{myteal}{\phantom{x}} achieve accuracy close to the full model ($88.4\%$).}
    \label{fig:pilot}
\end{figure}
\subsection{Isolating the Effects of Value-Awareness and Stochasticity}
\label{sec:ablation}
We perform a case study on GSM8K to isolate the individual contributions of value-awareness and stochasticity to the accuracy of Qwen3-4B.

\paragraph{Value Awareness.}
We experiment with different reservation budgets $\Vslots$ to evaluate the efficacy of our value-scoring function $\mathrm{Range}(\vv)$ (Eq.~\ref{eq:value-score}).
To isolate the effect of our value scoring, we implement \ourmethodattn without sampling.
Specifically, we fill the remaining budget with KV pairs ranked highest by SnapKV's scoring function $\Ssnap$ (Eq.~\ref{eq:snapkv-score}).
\Cref{fig:pilot} shows that when increasing the number of $\Vslots$ from 16 to 256 (green bars), the accuracies on GSM8K increase consistently from $73.3\% (+9.0\%)$ to $80.5\% (+16.2\%)$, showing the importance of keeping large-range values from eviction.

\paragraph{Stochasticity.}
We introduce stochasticity into SnapKV and CurDKV to promote diversity in KV pair retention.
For SnapKV (\textit{leftmost} in \Cref{fig:pilot}), replacing its deterministic top$k$ selection with $\bar\alpha_i$-weighted sampling improves model accuracy from $64.3\%$ to $70.9\%$.
Similarly, for CurDKV (\textit{rightmost}), resampling Gaussian matrix $\mG_t$ at each eviction step $t$ throughout generation improves accuracy from $78.6\%$ to $87.6\%$.
The teal bars \colorbox{myteal}{\phantom{x}} show that value scoring and stochasticity are complementary, with the combination of both strategies achieving the highest accuracies.

\subsection{Main Results}
\label{sec:main_exp}
\paragraph{Setup.}
We evaluate \ourmethod{} on Qwen3-4B and Qwen3-14B~\citep{yang2025qwen3} across six reasoning tasks: AIME25, AIME26 \citep{aime}, HMMT25 (Feb and Nov splits combined; \citealt{hmmt2025}), GPQA-Diamond \citep{rein2024gpqa}, MATH \citep{hendrycks2021math}, and LiveCodeBench-v6 \citep{jain2025livecodebench}.\footnote{We use the same evaluation scripts as SeerAttention-R: \url{https://github.com/microsoft/SeerAttention}} All eviction methods operate under the periodic-eviction framework (\S~\ref{sec:problem}) with a shared recency buffer of $B = 64$ and a persistent budget $K$. 
For each task, we first measure the average number of tokens $\Fullavg$ under the uncompressed full model and then set the eviction budget $K$ to approximately $\Fullavg / 4$, where $K \in \{1024, 2048, 4096\}$. This yields a nominal $4\times$ compression ratio over the uncompressed baseline. SeerAttention-R is configured at the matching $25\%$ activation ratio so that all sparse methods operate at the same effective attention sparsity.
We use the models' default top-p configurations for all generations.
We report pass@1 results, which are averaged over 16 runs for datasets under 100 examples, and 8 runs for larger datasets.
We select the hyperparameters on GSM8K for each method and apply them to other tasks without further tuning (see Appendix~\ref{sec:app_hyper}).

\begin{table}[tp]
  \centering
  \resizebox{\textwidth}{!}{
  \renewcommand{\arraystretch}{1.15}
  \small
  \begin{tabular}{l l c c c c c c}
  \toprule
  \textbf{Model} & \textbf{Method} &
  \textbf{AIME25} & \textbf{AIME26} & \textbf{HMMT25} &
  \textbf{GPQA-D} & \textbf{MATH} &
  \textbf{Avg.} \\
  \midrule

  % ────────────────────────────── Qwen3-4B ──────────────────────────────
  \multirow{7}{*}{\textbf{Qwen3-4B}}
    & \textcolor{black!65}{Full}
        & \textcolor{black!65}{66.41}
        & \textcolor{black!65}{62.71}
        & \textcolor{black!65}{45.94}
        & \textcolor{black!65}{56.19}
        & \textcolor{black!65}{93.93}
        & \textcolor{black!65}{65.04} \\
    \rowcolor{seerrow}
    \cellcolor{white} & {SeerAttention-R}
        & {58.59}
        & {60.42}
        & {40.42}
        & {49.94}
        & {84.67}
        & {58.81} \\
  %\cmidrule(lr){2-8}
    & SnapKV          & 47.92 & 53.33 & 35.31 & 36.87 & 72.30 & 49.15 \\
    & \KVRkv          & 49.58 & 54.58 & 36.98 & \textbf{49.05} & 83.28 & 54.69 \\
    & \KVCur          & 48.75 & 54.58 & 33.23 & 37.94 & 74.42 & 49.78 \\
    \rowcolor{ourrow}
    \cellcolor{white} & \ourmethoddkv
                    & 57.29 & 61.88 & 39.58 & 43.94 & \textbf{84.72} & 57.48 \\
    \rowcolor{ourrow}
    \cellcolor{white} & \ourmethodattn
                    & \textbf{59.17} & \textbf{62.08} & \textbf{44.38} & 47.98 & 81.82 & \textbf{59.09} \\
  \midrule

  % ────────────────────────────── Qwen3-14B ─────────────────────────────
  \multirow{6}{*}{\textbf{Qwen3-14B}}
    & \textcolor{black!65}{Full}
        & \textcolor{black!65}{70.21}
        & \textcolor{black!65}{76.04}
        & \textcolor{black!65}{54.79}
        & \textcolor{black!65}{65.25}
        & \textcolor{black!65}{95.22}
        & \textcolor{black!65}{72.30} \\
    \rowcolor{seerrow}
    \cellcolor{white} & {SeerAttention-R}
        & {64.79}
        & {65.62}
        & {48.65}
        & {61.68}
        & {86.12}
        & {65.37} \\
  %\cmidrule(lr){2-8}
    & \KVRkv          & 54.58 & 60.21 & 44.58 & \textbf{59.09} & 86.05 & 60.90 \\
    & \KVCur          & 53.75 & 61.25 & 39.58 & 48.48 & 75.42 & 55.70 \\
    \rowcolor{ourrow}
    \cellcolor{white} & \ourmethoddkv
                    & 63.33 & 68.75 & 49.38 & 56.76 & \textbf{86.47} & 64.94 \\
    \rowcolor{ourrow}
    \cellcolor{white} & \ourmethodattn
                    & \textbf{63.96} & \textbf{72.29} & \textbf{50.00} & 57.39 & 85.42 & \textbf{65.81} \\
  \bottomrule
  \end{tabular}
  }
  \vspace{1ex}
  \caption{%
  Reasoning-task accuracy (\%) of sparse attention methods on Qwen3-4B and
  Qwen3-14B with $\sim25\%$ of full KV activated ($4\times$ compression for eviction methods).
  \textbf{{Bold}} marks the best eviction method of each task. Our eviction methods \ourmethod\ achieve comparable average accuracy to SeerAttention-R, the
  SOTA selection method.
  }
  \label{tab:main_results}
  \end{table}

\paragraph{\ourmethodattn\ achieves the best average accuracy.}
\Cref{tab:main_results} shows that both \ourmethod{} variants outperform every prior eviction baseline on the per-model average and reach parity with the selection method \KVSeerR. 
On Qwen3-4B, \ourmethodattn{} achieves an average accuracy of $59.09\%$, edging out \KVSeerR\ ($58.81\%$) and surpassing the strongest eviction baseline R-KV ($54.69\%$) by $4.4\%$; \ourmethoddkv{} reaches $57.48\%$, improving over \KVCur\ ($49.78\%$) by $7.7\%$. 
The same pattern holds on Qwen3-14B: \ourmethodattn{} ($65.81\%$) matches \KVSeerR\ ($65.37\%$) and outperforms R-KV ($60.90\%$) by $4.9\%$; \ourmethoddkv{} ($64.94\%$) improves over the deterministic \KVCur\ baseline ($55.70\%$) by $9.2\%$.
Overall, \ourmethodattn{} achieves the best average accuracy among all methods.
While SnapKV and R-KV baselines do not consider value states in their deterministic scoring, the results of \ourmethodattn{} show the effectiveness of combining value awareness with stochasticity in the eviction scoring function.
The substantial improvements of \ourmethoddkv{} over \KVCur\ further highlight the importance of stochasticity.
Since both variants of \ourmethod{} belong to the eviction family, we demonstrate that the \ourmethod{} recipe can recover accuracy without paying the memory cost of selection methods.

\paragraph{Code generation.}
\Cref{fig:livecodebench} (\textit{Left}) reports pass@1 of Qwen3-4B on LiveCodeBench-v6-Medium under a 2048-token budget ($\sim 20\%$ of full KV size).
R-KV ($62.6\%$), \ourmethoddkv{} ($61.9\%$), and \ourmethodattn{} ($63.5\%$) achieve comparable performance, while CurDKV only yields $34.6\%$.
Surprisingly, SeerAttention-R ($45.3\%$) underperforms eviction methods, with the exception of CurDKV.
This is likely because the learning-based attention gate of SeerAttention-R struggles to generalize under domain shifts, while the training-free eviction methods are less affected.

\begin{figure}[t]
    \centering
    \includegraphics[width=\linewidth]{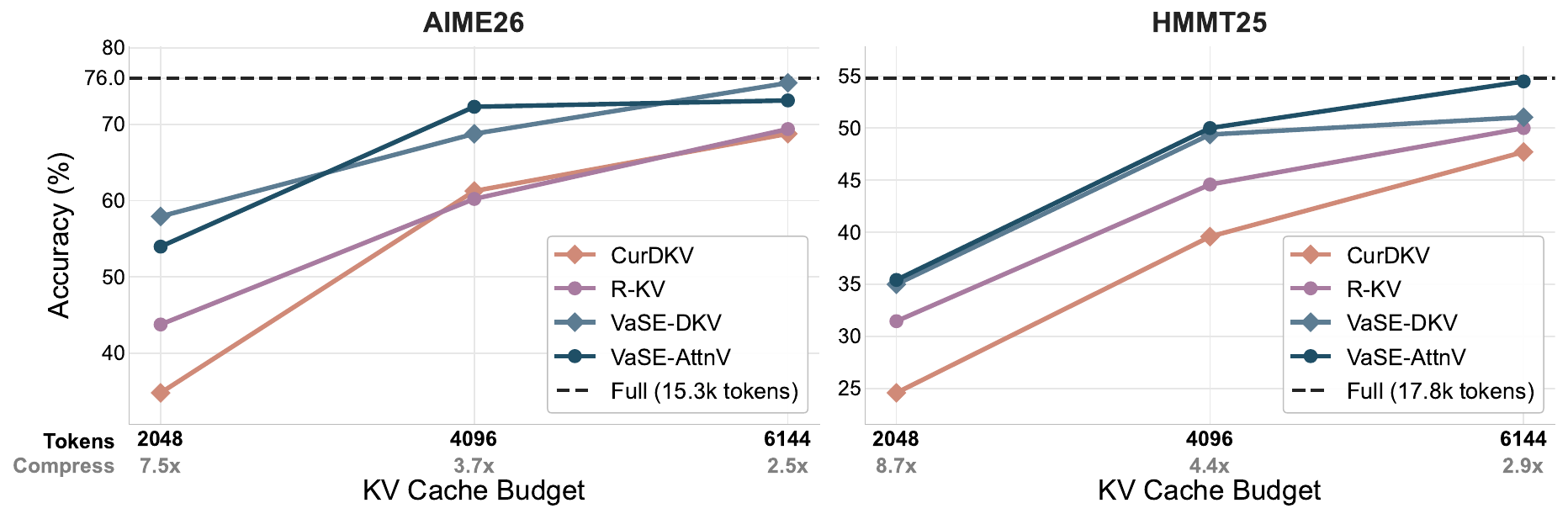}
    \caption{Pass@1 accuracy of Qwen3-14B under varying KV cache budgets. The full-cache model uses an average of 15.3k and 17.8k tokens on AIME26 and HMMT25, respectively. Each x-axis label reports the number of budget tokens (e.g., 2048) and its corresponding KV cache compression ratio to the full model (e.g., 7.5$\times$).}
    \label{fig:budget}
\end{figure}

\textbf{Token Budgets.} Figure~\ref{fig:budget} reports pass@1 accuracy of Qwen3-14B on AIME26 and HMMT25 as the KV cache budget increases from 2048 to 6144 tokens. At the tightest budget of 2048 tokens, which corresponds to roughly $7.5\times$ compression of the full cache on AIME26 and $8.7\times$ on HMMT25, both \ourmethod{} variants show clear improvement over R-KV and CurDKV baselines.
%as they both outperform the best baseline by up to 17.92\% and 8.54\% on AIME26, and 9.38\% and 6.46\% on HMMT25, respectively.
%\dfcomment{TODO: quote exact accuracy values from figure for each method at 2048}. 
%As the budget grows, all methods recover accuracy, and the gap between them narrows, while \ourmethod{} methods still outperform baselines (green and teal lines).
As the budget increases, the performance gap diminishes with the accuracy of all methods recovering.
Nevertheless, \ourmethod{} methods (teal lines) consistently outperform the baselines.
%with all variants converging toward the full cache baselines of 76.0\% and 54.8\% at 6144 tokens. \dfcomment{TODO: confirm full-cache numbers from figure}. 
We observe this pattern in both datasets: the accuracy gains from value awareness and stochastic eviction are largest under aggressive compression and diminish gracefully when the cache capacity is abundant (see \Cref{tab:gpqa_budgets} in the appendix for results on GPQA-Diamond).

\subsection{Large-\textbf{Range} Values Correlate with KV Cache Quantization Error}
\label{sec:quantization}
Our findings regarding large-$\Range$ values have a deep connection with per-token KV cache quantization \citep{liu2024kivi,su2025kvsink}, an alternative way to compress the KV cache.
Formally, in asymmetric $b$-bit linear quantization, the range of the input value state $\vv$, $[\min \vv, \max \vv]$, is mapped to the full range of the quantized integer space, $[0, 2^b- 1]$. 
The $b$-bit integer quantization-dequantization process can be expressed as:
\begin{equation}
Q(\vv) = \left\lfloor \frac{\vv - z_\vv}{\Scale} \right\rceil, \quad \vv^\prime = Q(\vv) \cdot \Scale + z_\vv,
\end{equation}
where $z_\vv = \min \vv$ is the zero point, $\Scale = (\max \vv - \min \vv) / (2^b - 1) = \Range(\vv) / (2^b - 1)$ is the scaling factor, $\left\lfloor \cdot \right\rceil$ is the rounding operation, and $\vv^\prime$ is the reconstructed vector of $\vv$ after dequantization.
The scaling factor $\Scale$ represents the step size of quantization: if $\Range(\vv)$ is small, $\Scale$ is small, leading to fine-grained quantization, and the reconstructed value $\vv^\prime$ is close to the original value $\vv$; on the other hand, if $\Range(\vv)$ is large, a large $\Scale$ results in a large and coarse step size, causing $\vv^\prime$ to deviate significantly from $\vv$, which is known as the outlier issue \citep{dettmers2022gptint} in LLM quantization.

\begin{figure}[tp]
    \centering
    \includegraphics[width=.85\linewidth]{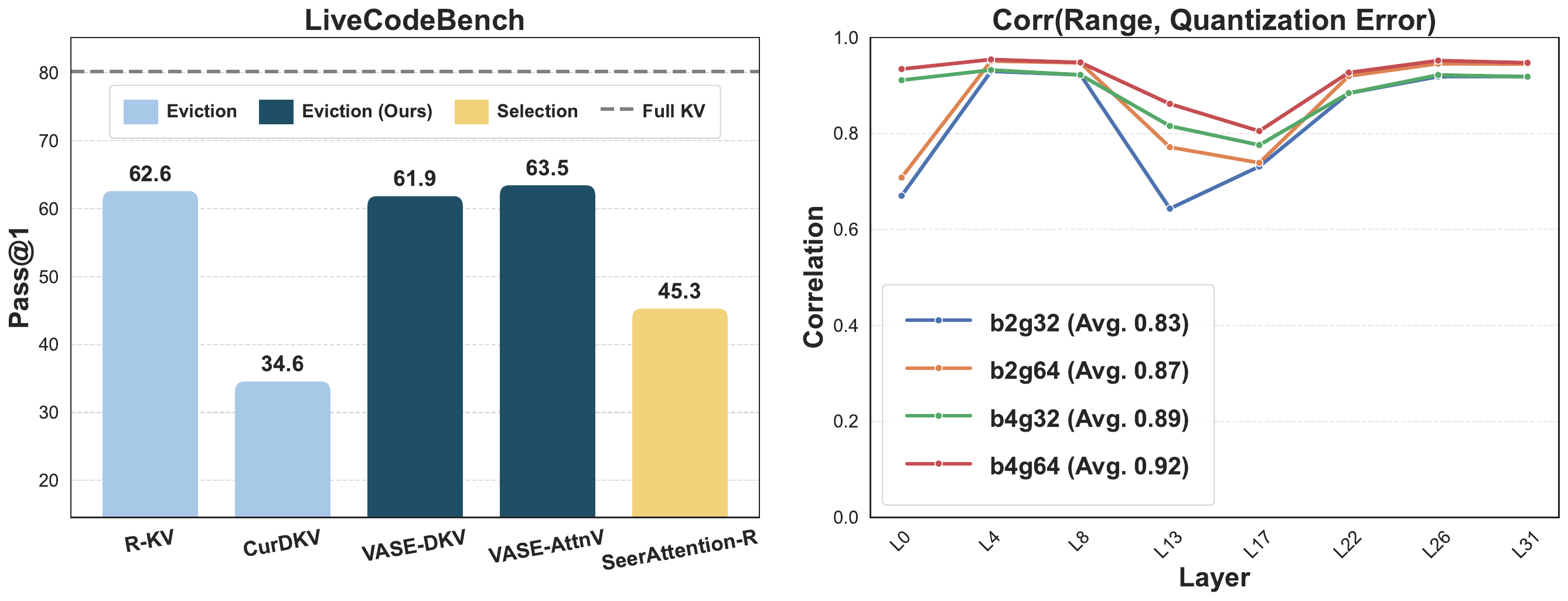}
    \caption{\textit{Left}: Pass@1 results of Qwen3-4B on LiveCodeBench under a 2048-token budget (${\sim}20\%$ of full KV). R-KV and our \ourmethod\ methods achieve the strongest performance. \textit{Right}: $\Range(\vv)$ and per-token value cache quantization errors are highly correlated under different quantization configurations, where \texttt{b2g32} means 2-bit precision with a group size of 32.}
    %Surprisingly, the selection-based method \KVSeerR\ performs poorly, likely due to the absence of code-related data during the training of its attention gates, causing train-test mismatch.
    \label{fig:livecodebench}
\end{figure}

We validate the relationship between $\Range(\vv)$ and per-token quantization error on the value cache empirically, using the HQQ quantizer \citep{badri2023hqq} with different bit-widths and quantization group sizes.
Quantization error is measured via the mean-squared reconstruction error between $\vv$ and $\vv^\prime$ \citep{frantar2023optq}. 
We extract the full value cache from Qwen3-4B during the generation of GSM8K examples, quantize them, and compute the per-token mean-squared error.
\Cref{fig:livecodebench} (\textit{Right}) demonstrates that $\Range(\vv)$ and the reconstruction error are highly correlated across layers under different quantization configurations, suggesting that large-$\Range$ value states lead to large information loss under per-token KV cache quantization.

\begin{figure}[t]
    \centering
    \includegraphics[width=\linewidth]{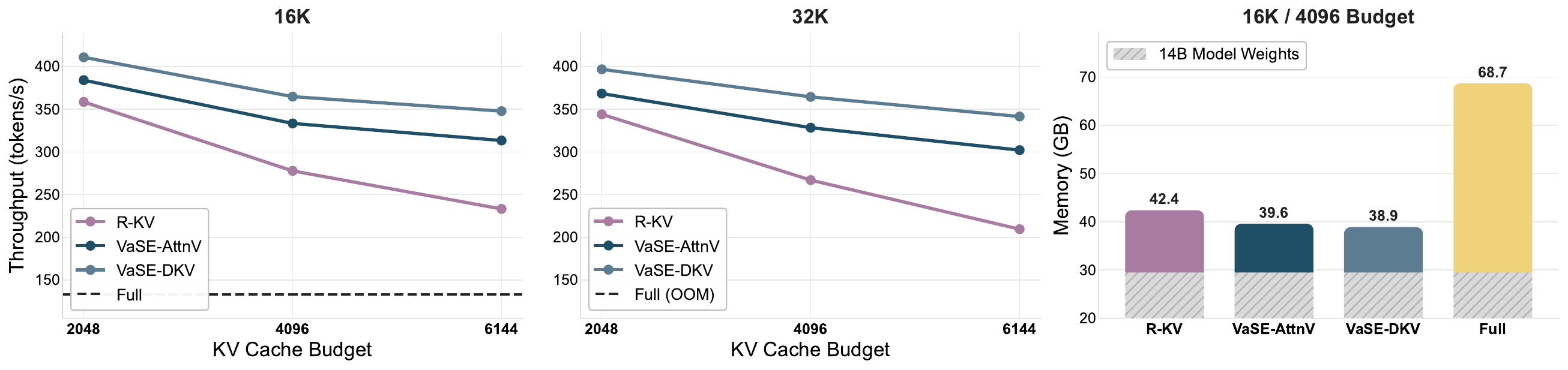}
    \caption{\textit{Left \& Middle}: Decode throughput ($\uparrow$) of the Qwen3-14B model on a single A100-80G GPU under different KV cache budgets $\{$2048, 4096, 6144$\}$ and total output tokens $\{$16K, 32K$\}$. All eviction methods run well above the original \KVFull\ method (dashed line; OOM at 32K), with \ourmethoddkv\ achieving the fastest throughput. \textit{Right}: Peak GPU memory ($\downarrow$) at the 16K output tokens and 4096 budget; the 14B model weights (hatched) account for nearly 30GB of memory. All eviction methods have a much lower memory footprint than \KVFull, with \ourmethoddkv\ using the least memory.}
    \label{fig:thpt}
\end{figure}
  \section{Benchmark Throughput and Memory}
  \label{sec:benchmark}                                                               
  In this section, we benchmark the actual decode-phase throughput and peak memory of different methods on the Qwen3-14B model using a single A100-80G GPU.
  We sweep over KV cache budgets $\{2048, 4096, 6144\}$ and total output tokens $\{16384, 32768\}$.
  All methods are benchmarked with FlashAttention2 kernels \citep{dao2023flashattention2}, without PagedAttention \citep{pagedattn}.
  Our benchmark script builds on that of \citet{song2025reasoning}; we set the batch size to $16$ and the input prompt length to $256$ tokens across all experiments.

  \Cref{fig:thpt} (\textit{Left \& Middle}) shows that all eviction methods achieve substantially higher throughput than the \KVFull\ baseline, with a consistent
  ordering of \ourmethoddkv $>$ \ourmethodattn $>$ \KVRkv $>$ \KVFull\ across all settings.
  For example, at 16K output tokens and a KV cache budget of 2048, \ourmethoddkv\ is $3.1\times$ faster than \KVFull\ ($411$ vs.\ $133$ tokens per second).
  Our \ourmethoddkv\ is the fastest eviction method because it does not compute attention scores during eviction, and \ourmethodattn\ outperforms \KVRkv\ because it
  avoids the additional redundancy-score computation used to promote diversity.
  For all eviction methods, the throughput decreases with increase of the token budget.
  On the other hand, the ordering of memory footprint (\Cref{fig:thpt}, \textit{Right}, $\downarrow$) is reversed: \ourmethoddkv $<$ \ourmethodattn $<$ \KVRkv $<$ \KVFull.
  \Cref{fig:memory} in the appendix shows consistent ranking under different token budgets.
  Excluding the memory of the 14B model weights (the hatched bars), our \ourmethod\ methods achieve roughly $4\times$ theoretical KV cache compression (16K/4K) over the \KVFull\ baseline.
\section{Related Work} \label{sec:related}
\paragraph{KV Cache Compression Methods.}
%In autoregressive generation, Transformer models \citep{vaswani2017attention} cache every previously generated key and value state, allowing the model to attend to the representations of the context at each decode step.
%As a result, the memory of the KV cache grows linearly with the number of tokens.
Prior work reduces the cost of the KV cache through different approaches: reducing the number of tokens via sparse attention \citep{liu2023scissorhands,jiang2024minference,singhania2024loki,chari2025compactor}, reducing the precision via quantization \citep{hooperkvquant,liu2024kivi,kim2025not,zandieh2026turboquant}, and reducing the hidden dimension via low-rank decomposition \citep{saxena2024eigen,liu2024deepseek,chang2025palu}.
In this paper, we focus on sparse attention methods.
%, showing that our proposed recipe for eviction-based methods achieves accuracy on par with selection-based methods while substantially reducing KV cache memory. 
Our findings on the importance of large-magnitude values also carry profound implications for per-token KV cache quantization \citep{liu2024kivi}.

\paragraph{Decoding Phase Compression.} Inference proceeds in two phases.
In the prefill phase, the model processes the $T$-token input prompt and computes $T$ key--value pairs in parallel.
In the decode phase, the model generates one token at a time and appends the corresponding KV pairs to the cache.
Unlike the prefill phase, the decode phase is memory-bound, and the growing cache intensifies the memory bottleneck.
For reasoning models, the KV pairs from the decode phase dominate the cost due to the long thinking traces.
%: models may generate over 10{,}000 tokens at decode steps for a math question that has fewer than 200 prefill tokens (see Table~\ref{tab:dataset_tokens} for the statistics).
Therefore, several recent works \citep{cai2025rkv,song2025reasoning,gao2026sparse,guo2026dynamic} have targeted KV cache compression during the decoding phase of reasoning models.
Our paper also focuses on the decoding phase compression of reasoning models.
Unlike \citet{song2025reasoning}, we do not distinguish between prompt tokens and generated tokens at the eviction step; therefore, prompt tokens may also be evicted in our experiments.

\paragraph{Leveraging Value-State Magnitude for KV Cache Scoring.}
Typically, sparse attention involves scoring key-value pairs by their importance. %and pruning those with the lowest scores.
%For example, H2O \citep{zhang2023h2o} and SnapKV \citep{li2024snapkv} evict KV pairs with low average attention scores. 
Because the output of attention is a weighted combination of values, \citet{guo-etal-2024-attention,devoto2025expected} score KV pairs by weighting attention scores with the norm of their corresponding values.
In this paper, we deeply investigate the importance of large-magnitude values.
Rather than multiplying attention and value scores, we propose an alternative, \ourmethodattn{}, which reserves dedicated slots in the token budget to ensure that large-magnitude values are always preserved in the cache.

%stocastic
\section{Discussion and Conclusion} \label{sec:discussion}

\paragraph{Implications for KV cache quantization.}
We show that large-$\Range$ value states are not only critical for maintaining KV cache eviction accuracy but also a major source of error under quantization.
%We show that large-$\Range$ values are not only critical for maintaining KV cache eviction accuracy, but also induce large errors under quantization.
Therefore, a promising future direction is to investigate a mixed-precision approach \citep{liu2024kivi} to maintain the precision of large-$\Range$ value states. 
For example, they can be placed in a high-precision reserved cache, while the remaining KV cache is quantized to a lower bit-width. Once the high-precision budget is used up, the earliest value state in this reserved cache can be quantized and ``evicted'' into the low-precision cache.

\paragraph{Large-magnitude value states are crucial for reasoning progression.}
Our observation in Figure \ref{fig:loop} suggests that large-magnitude value states play a special role in maintaining reasoning progression. After these values are evicted, generations often degenerate into repetitive loops, where the model repeatedly restates intermediate reasoning without making progress. One hypothesis is that these values help transition between latent reasoning steps, preventing the model from collapsing into self-reinforcing paths.
%help mediate transitions between reasoning stages, allowing the model to move away from locally self-reinforcing continuations. 
%Their removal can make the model trapped in a repetitive region of the generation trajectory.
Therefore, their removal traps the model in repetitive loops.
One potential future direction is to analyze what information large-magnitude value states carry and how they influence long-form reasoning.
\paragraph{Conclusion.} \label{sec:conclusion}
We present \ourmethod{}, a training-free KV cache eviction framework designed to address the memory bottleneck of reasoning models. We identify two critical factors for maintaining accuracy during KV cache compression: protecting large-magnitude value states from eviction, and introducing stochasticity to improve KV cache diversity. Taking both factors into consideration, \ourmethod{} provides a simple and effective recipe for improving existing eviction methods. More broadly, our findings highlight the importance of value states in long-form reasoning and suggest new directions for designing memory-efficient inference methods.

%\clearpage
\makeatletter
\if@preprint
\section*{Acknowledgment}
This work was supported in part by
a gift from the Capital One Center for Responsible AI and Decision Making in Finance (CREDIF), and by the National
Science Foundation under Grant No. IIS-2403436. Any opinions, findings, and conclusions or recommendations expressed in this material are those of the author(s) and do not necessarily reflect the
views of the National Science Foundation.

\fi 
\makeatother
\clearpage
\bibliographystyle{plainnat}
\bibliography{main}

@article{zhang2023h2o,
  title={H2o: Heavy-hitter oracle for efficient generative inference of large language models},
  author={Zhang, Zhenyu and Sheng, Ying and Zhou, Tianyi and Chen, Tianlong and Zheng, Lianmin and Cai, Ruisi and Song, Zhao and Tian, Yuandong and R{\'e}, Christopher and Barrett, Clark and others},
  journal={Advances in Neural Information Processing Systems},
  volume={36},
  pages={34661--34710},
  year={2023}
}

@inproceedings{
xiao2024efficient,
title={Efficient Streaming Language Models with Attention Sinks},
author={Guangxuan Xiao and Yuandong Tian and Beidi Chen and Song Han and Mike Lewis},
booktitle={The Twelfth International Conference on Learning Representations},
year={2024},
}

@inproceedings{
li2024snapkv,
title={Snap{KV}: {LLM} Knows What You are Looking for Before Generation},
author={Yuhong Li and Yingbing Huang and Bowen Yang and Bharat Venkitesh and Acyr Locatelli and Hanchen Ye and Tianle Cai and Patrick Lewis and Deming Chen},
booktitle={The Thirty-eighth Annual Conference on Neural Information Processing Systems},
year={2024},
}

@inproceedings{
jiang2024minference,
title={{MI}nference 1.0: Accelerating Pre-filling for Long-Context {LLM}s via Dynamic Sparse Attention},
author={Huiqiang Jiang and YUCHENG LI and Chengruidong Zhang and Qianhui Wu and Xufang Luo and Surin Ahn and Zhenhua Han and Amir H. Abdi and Dongsheng Li and Chin-Yew Lin and Yuqing Yang and Lili Qiu},
booktitle={The Thirty-eighth Annual Conference on Neural Information Processing Systems},
year={2024},
}

@inproceedings{oren-etal-2024-transformers,
    title = "Transformers are Multi-State {RNN}s",
    author = "Oren, Matanel  and
      Hassid, Michael  and
      Yarden, Nir  and
      Adi, Yossi  and
      Schwartz, Roy",
    booktitle = "Proceedings of the 2024 Conference on Empirical Methods in Natural Language Processing",
    month = nov,
    year = "2024",
    address = "Miami, Florida, USA",
    publisher = "Association for Computational Linguistics",
    url = "https://aclanthology.org/2024.emnlp-main.1043/",
    doi = "10.18653/v1/2024.emnlp-main.1043",
    pages = "18724--18741"
}

@inproceedings{
gao2026sparse,
title={Sparse Attention Adaptation for Long Reasoning},
author={Yizhao Gao and Shuming Guo and Shijie Cao and Yuqing Xia and Yu Cheng and Lei Wang and Lingxiao Ma and Yutao Sun and Tianzhu Ye and Li Dong and Hayden Kwok-Hay So and Yu Hua and Ting Cao and Fan Yang and Mao Yang},
booktitle={The Fourteenth International Conference on Learning Representations},
year={2026},
url={https://openreview.net/forum?id=c5BOcHM6J8}
}

@inproceedings{ribarsparq,
  title = 	 {{S}par{Q} Attention: Bandwidth-Efficient {LLM} Inference},
  author =       {Ribar, Luka and Chelombiev, Ivan and Hudlass-Galley, Luke and Blake, Charlie and Luschi, Carlo and Orr, Douglas},
  booktitle = 	 {Proceedings of the 41st International Conference on Machine Learning},
  pages = 	 {42558--42583},
  year = 	 {2024},
  editor = 	 {Salakhutdinov, Ruslan and Kolter, Zico and Heller, Katherine and Weller, Adrian and Oliver, Nuria and Scarlett, Jonathan and Berkenkamp, Felix},
  volume = 	 {235},
  series = 	 {Proceedings of Machine Learning Research},
  month = 	 {21--27 Jul},
  publisher =    {PMLR},
  url = 	 {https://proceedings.mlr.press/v235/ribar24a.html},
}

@inproceedings{tang2024quest,
  title={QUEST: Query-Aware Sparsity for Efficient Long-Context LLM Inference},
  author={Tang, Jiaming and Zhao, Yilong and Zhu, Kan and Xiao, Guangxuan and Kasikci, Baris and Han, Song},
  booktitle = {Proceedings of the 41st International Conference on Machine Learning},
  year      = {2024}
}

@inproceedings{
yang2025tidaldecode,
title={TidalDecode: Fast and Accurate {LLM} Decoding with Position Persistent Sparse Attention},
author={Lijie Yang and Zhihao Zhang and Zhuofu Chen and Zikun Li and Zhihao Jia},
booktitle={The Thirteenth International Conference on Learning Representations},
year={2025},
url={https://openreview.net/forum?id=EkfLaCJ7bk}
}

@inproceedings{
cai2025rkv,
title={R-{KV}: Redundancy-aware {KV} Cache Compression for Reasoning Models},
author={Zefan Cai and Wen Xiao and Hanshi Sun and Cheng Luo and Yikai Zhang and Ke Wan and Yucheng Li and Yeyang Zhou and Li-Wen Chang and Jiuxiang Gu and Zhen Dong and Anima Anandkumar and Abedelkadir Asi and Junjie Hu},
booktitle={The Thirty-ninth Annual Conference on Neural Information Processing Systems},
year={2025},
}

@inproceedings{
song2025reasoning,
title={Reasoning Path Compression: Compressing Generation Trajectories for Efficient {LLM} Reasoning},
author={Jiwon Song and Dongwon Jo and Yulhwa Kim and Jae-Joon Kim},
booktitle={The Thirty-ninth Annual Conference on Neural Information Processing Systems},
year={2025},
}

@inproceedings{
chen2024not,
title={Do {NOT} Think That Much for 2+3=? On the Overthinking of Long Reasoning Models},
author={Xingyu Chen and Jiahao Xu and Tian Liang and Zhiwei He and Jianhui Pang and Dian Yu and Linfeng Song and Qiuzhi Liu and Mengfei Zhou and Zhuosheng Zhang and Rui Wang and Zhaopeng Tu and Haitao Mi and Dong Yu},
booktitle={Forty-second International Conference on Machine Learning},
year={2025},
url={https://openreview.net/forum?id=MSbU3L7V00}
}

@misc{openai2024o1,
    author       = {OpenAI},
    title        = {Learning to Reason with {LLM}s},
    year         = {2024},
    howpublished = {\url{https://openai.com/index/learning-to-reason-with-llms}},
  }

@article{guo2025deepseek,
  title={Deepseek-r1: Incentivizing reasoning capability in llms via reinforcement learning},
  author={Guo, Daya and Yang, Dejian and Zhang, Haowei and Song, Junxiao and Wang, Peiyi and Zhu, Qihao and Xu, Runxin and Zhang, Ruoyu and Ma, Shirong and Bi, Xiao and others},
  journal={arXiv preprint arXiv:2501.12948},
  year={2025}
}

@misc{qwen2024qwq,
    author       = {Qwen},
    title        = {{QwQ}: Reflect Deeply on the Boundaries of the Unknown},
    year         = {2024},
    month        = nov,
    howpublished = {\url{https://qwenlm.github.io/blog/qwq-32b-preview/}},
  }

@inproceedings{
snell2025scaling,
title={Scaling {LLM} Test-Time Compute Optimally Can be More Effective than Scaling Parameters for Reasoning},
author={Charlie Victor Snell and Jaehoon Lee and Kelvin Xu and Aviral Kumar},
booktitle={The Thirteenth International Conference on Learning Representations},
year={2025},
url={https://openreview.net/forum?id=4FWAwZtd2n}
}

@inproceedings{
sengupta2025cur,
title={Value-Guided {KV} Compression for {LLM}s via Approximated {CUR} Decomposition},
author={Ayan Sengupta and Siddhant Chaudhary and Tanmoy Chakraborty},
booktitle={The Thirty-ninth Annual Conference on Neural Information Processing Systems},
year={2025},
}

@article{beltagy2020longformer,
  title={Longformer: The long-document transformer},
  author={Beltagy, Iz and Peters, Matthew E and Cohan, Arman},
  journal={arXiv preprint arXiv:2004.05150},
  year={2020}
}

@article{zaheer2020big,
  title={Big bird: Transformers for longer sequences},
  author={Zaheer, Manzil and Guruganesh, Guru and Dubey, Kumar Avinava and Ainslie, Joshua and Alberti, Chris and Ontanon, Santiago and Pham, Philip and Ravula, Anirudh and Wang, Qifan and Yang, Li and others},
  journal={Advances in neural information processing systems},
  volume={33},
  pages={17283--17297},
  year={2020}
}

@inproceedings{
kitaevreformer,
title={Reformer: The Efficient Transformer},
author={Nikita Kitaev and Lukasz Kaiser and Anselm Levskaya},
booktitle={International Conference on Learning Representations},
year={2020},
url={https://openreview.net/forum?id=rkgNKkHtvB}
}

@inproceedings{
gemodel,
title={Model Tells You What to Discard: Adaptive {KV} Cache Compression for {LLM}s},
author={Suyu Ge and Yunan Zhang and Liyuan Liu and Minjia Zhang and Jiawei Han and Jianfeng Gao},
booktitle={The Twelfth International Conference on Learning Representations},
year={2024},
url={https://openreview.net/forum?id=uNrFpDPMyo}
}

@article{zhao2019explicit,
  title={Explicit sparse transformer: Concentrated attention through explicit selection},
  author={Zhao, Guangxiang and Lin, Junyang and Zhang, Zhiyuan and Ren, Xuancheng and Su, Qi and Sun, Xu},
  journal={arXiv preprint arXiv:1912.11637},
  year={2019}
}

@inproceedings{devoto-etal-2024-simple,
    title = "A Simple and Effective $L\_2$ Norm-Based Strategy for {KV} Cache Compression",
    author = "Devoto, Alessio  and
      Zhao, Yu  and
      Scardapane, Simone  and
      Minervini, Pasquale",
    booktitle = "Proceedings of the 2024 Conference on Empirical Methods in Natural Language Processing",
    month = nov,
    year = "2024",
    address = "Miami, Florida, USA",
    publisher = "Association for Computational Linguistics",
    url = "https://aclanthology.org/2024.emnlp-main.1027/",
    doi = "10.18653/v1/2024.emnlp-main.1027",
    pages = "18476--18499",
}

@article{yang2025qwen3,
  title={Qwen3 technical report},
  author={Yang, An and Li, Anfeng and Yang, Baosong and Zhang, Beichen and Hui, Binyuan and Zheng, Bo and Yu, Bowen and Gao, Chang and Huang, Chengen and Lv, Chenxu and others},
  journal={arXiv preprint arXiv:2505.09388},
  year={2025}
}

@article{vaswani2017attention,
  title={Attention is all you need},
  author={Vaswani, Ashish and Shazeer, Noam and Parmar, Niki and Uszkoreit, Jakob and Jones, Llion and Gomez, Aidan N and Kaiser, {\L}ukasz and Polosukhin, Illia},
  journal={Advances in neural information processing systems},
  volume={30},
  year={2017}
}

@inproceedings{
hooperkvquant,
title={{KVQ}uant: Towards 10 Million Context Length {LLM} Inference with {KV} Cache Quantization},
author={Coleman Richard Charles Hooper and Sehoon Kim and Hiva Mohammadzadeh and Michael W. Mahoney and Sophia Shao and Kurt Keutzer and Amir Gholami},
booktitle={The Thirty-eighth Annual Conference on Neural Information Processing Systems},
year={2024},
url={https://openreview.net/forum?id=0LXotew9Du}
}

@inproceedings{liu2024kivi,
  title={KIVI: A Tuning-Free Asymmetric 2bit Quantization for KV Cache},
  author={Liu, Zirui and Yuan, Jiayi and Jin, Hongye and Zhong, Shaochen and Xu, Zhaozhuo and Braverman, Vladimir and Chen, Beidi and Hu, Xia},
  booktitle={International Conference on Machine Learning},
  pages={32332--32344},
  year={2024},
  organization={PMLR}
}

@article{kim2025not,
  title={Not All Bits Are Equal: Scale-Dependent Memory Optimization Strategies for Reasoning Models},
  author={Kim, Junhyuck and Ewer, Ethan and Moon, Taehong and Park, Jongho and Papailiopoulos, Dimitris},
  journal={arXiv preprint arXiv:2510.10964},
  year={2025}
}

@inproceedings{
zandieh2026turboquant,
title={TurboQuant: Online Vector Quantization with Near-optimal Distortion Rate},
author={Amir Zandieh and Majid Daliri and Majid Hadian and Vahab Mirrokni},
booktitle={The Fourteenth International Conference on Learning Representations},
year={2026},
url={https://openreview.net/forum?id=tO3ASKZlok}
}

@inproceedings{
liu2023scissorhands,
title={Scissorhands: Exploiting the Persistence of Importance Hypothesis for {LLM} {KV} Cache Compression at Test Time},
author={Zichang Liu and Aditya Desai and Fangshuo Liao and Weitao Wang and Victor Xie and Zhaozhuo Xu and Anastasios Kyrillidis and Anshumali Shrivastava},
booktitle={Thirty-seventh Conference on Neural Information Processing Systems},
year={2023},
url={https://openreview.net/forum?id=JZfg6wGi6g}
}

@inproceedings{saxena2024eigen,
  title={Eigen attention: Attention in low-rank space for kv cache compression},
  author={Saxena, Utkarsh and Saha, Gobinda and Choudhary, Sakshi and Roy, Kaushik},
  booktitle={Findings of the Association for Computational Linguistics: EMNLP 2024},
  pages={15332--15344},
  year={2024}
}

@inproceedings{
chang2025palu,
title={Palu: {KV}-Cache Compression with Low-Rank Projection},
author={Chi-Chih Chang and Wei-Cheng Lin and Chien-Yu Lin and Chong-Yan Chen and Yu-Fang Hu and Pei-Shuo Wang and Ning-Chi Huang and Luis Ceze and Mohamed S. Abdelfattah and Kai-Chiang Wu},
booktitle={The Thirteenth International Conference on Learning Representations},
year={2025},
url={https://openreview.net/forum?id=LWMS4pk2vK}
}

@article{liu2024deepseek,
  title={Deepseek-v2: A strong, economical, and efficient mixture-of-experts language model},
  author={Liu, Aixin and Feng, Bei and Wang, Bin and Wang, Bingxuan and Liu, Bo and Zhao, Chenggang and Dengr, Chengqi and Ruan, Chong and Dai, Damai and Guo, Daya and others},
  journal={arXiv preprint arXiv:2405.04434},
  year={2024}
}

@inproceedings{
singhania2024loki,
title={Loki: Low-rank Keys for Efficient Sparse Attention},
author={Prajwal Singhania and Siddharth Singh and Shwai He and Soheil Feizi and Abhinav Bhatele},
booktitle={The Thirty-eighth Annual Conference on Neural Information Processing Systems},
year={2024},
url={https://openreview.net/forum?id=raABeiV71j}
}

@article{devoto2025expected,
  title={Expected attention: Kv cache compression by estimating attention from future queries distribution},
  author={Devoto, Alessio and Jeblick, Maximilian and J{\'e}gou, Simon},
  journal={arXiv preprint arXiv:2510.00636},
  year={2025}
}

@article{chari2025compactor,
  title={Compactor: Calibrated Query-Agnostic KV Cache Compression with Approximate Leverage Scores},
  author={Chari, Vivek and Van Durme, Benjamin},
  journal={arXiv preprint arXiv:2507.08143},
  year={2025}
}

@inproceedings{guo-etal-2024-attention,
    title = "Attention Score is not All You Need for Token Importance Indicator in {KV} Cache Reduction: Value Also Matters",
    author = "Guo, Zhiyu  and
      Kamigaito, Hidetaka  and
      Watanabe, Taro",
    booktitle = "Proceedings of the 2024 Conference on Empirical Methods in Natural Language Processing",
    month = nov,
    year = "2024",
    address = "Miami, Florida, USA",
    publisher = "Association for Computational Linguistics",
    url = "https://aclanthology.org/2024.emnlp-main.1178/",
    doi = "10.18653/v1/2024.emnlp-main.1178",
    pages = "21158--21166"
}

@inproceedings{
guoactive,
title={Active-Dormant Attention Heads: Mechanistically Demystifying Extreme-Token Phenomena in {LLM}s},
author={Tianyu Guo and Druv Pai and Yu Bai and Jiantao Jiao and Michael I. Jordan and Song Mei},
booktitle={The Second Conference on Parsimony and Learning (Recent Spotlight Track)},
year={2025},
url={https://openreview.net/forum?id=Zx6WUbE9J7}
}

@misc{aime,
    author       = {{Art of Problem Solving}},
    title        = {{AIME} Problems and Solutions},
    howpublished = {\url{https://artofproblemsolving.com/wiki/index.php/AIME_Problems_and_Solutions}}
  }

@misc{hmmt2025,
    author       = {{HMMT}},
    title        = {{HMMT} 2025},
    year         = {2025},
    url          = {https://www.hmmt.org/},
    note         = {Accessed: 2025}
  }

@inproceedings{
hendrycks2021math,
title={Measuring Mathematical Problem Solving With the {MATH} Dataset},
author={Dan Hendrycks and Collin Burns and Saurav Kadavath and Akul Arora and Steven Basart and Eric Tang and Dawn Song and Jacob Steinhardt},
booktitle={Thirty-fifth Conference on Neural Information Processing Systems Datasets and Benchmarks Track (Round 2)},
year={2021},
url={https://openreview.net/forum?id=7Bywt2mQsCe}
}

@inproceedings{
jain2025livecodebench,
title={LiveCodeBench: Holistic and Contamination Free Evaluation of Large Language Models for Code},
author={Naman Jain and King Han and Alex Gu and Wen-Ding Li and Fanjia Yan and Tianjun Zhang and Sida Wang and Armando Solar-Lezama and Koushik Sen and Ion Stoica},
booktitle={The Thirteenth International Conference on Learning Representations},
year={2025},
url={https://openreview.net/forum?id=chfJJYC3iL}
}

@inproceedings{
rein2024gpqa,
title={{GPQA}: A Graduate-Level Google-Proof Q\&A Benchmark},
author={David Rein and Betty Li Hou and Asa Cooper Stickland and Jackson Petty and Richard Yuanzhe Pang and Julien Dirani and Julian Michael and Samuel R. Bowman},
booktitle={First Conference on Language Modeling},
year={2024},
url={https://openreview.net/forum?id=Ti67584b98}
}

@article{cobbe2021gsm8k,
  title={Training Verifiers to Solve Math Word Problems},
  author={Cobbe, Karl and Kosaraju, Vineet and Bavarian, Mohammad and Chen, Mark and Jun, Heewoo and Kaiser, Lukasz and Plappert, Matthias and Tworek, Jerry and Hilton, Jacob and Nakano, Reiichiro and Hesse, Christopher and Schulman, John},
  journal={arXiv preprint arXiv:2110.14168},
  year={2021}
}

@inproceedings{
dettmers2022gptint,
title={{GPT}3.int8(): 8-bit Matrix Multiplication for Transformers at Scale},
author={Tim Dettmers and Mike Lewis and Younes Belkada and Luke Zettlemoyer},
booktitle={Advances in Neural Information Processing Systems},
editor={Alice H. Oh and Alekh Agarwal and Danielle Belgrave and Kyunghyun Cho},
year={2022},
url={https://openreview.net/forum?id=dXiGWqBoxaD}
}

@inproceedings{
frantar2023optq,
title={{OPTQ}: Accurate Quantization for Generative Pre-trained Transformers},
author={Elias Frantar and Saleh Ashkboos and Torsten Hoefler and Dan Alistarh},
booktitle={The Eleventh International Conference on Learning Representations },
year={2023},
url={https://openreview.net/forum?id=tcbBPnfwxS}
}

@inproceedings{
su2025kvsink,
title={{KVS}ink: Understanding and Enhancing the Preservation of Attention Sinks in {KV} Cache Quantization for {LLM}s},
author={Zunhai Su and Kehong Yuan},
booktitle={Second Conference on Language Modeling},
year={2025},
url={https://openreview.net/forum?id=gIqb6zWZoO}
}

@misc{badri2023hqq,
title  = {Half-Quadratic Quantization of Large Machine Learning Models},
url    = {https://dropbox.github.io/hqq_blog/},
author = {Hicham Badri and Appu Shaji},
month  = {November},
year   = {2023}
}

@article{guo2026dynamic,
  title={Dynamic Thinking-Token Selection for Efficient Reasoning in Large Reasoning Models},
  author={Guo, Zhenyuan and Chen, Tong and Meng, Wenlong and Gong, Chen and Yu, Xin and Wei, Chengkun and Chen, Wenzhi},
  journal={arXiv preprint arXiv:2601.18383},
  year={2026}
}

@inproceedings{dao2023flashattention2,
  title={Flash{A}ttention-2: Faster Attention with Better Parallelism and Work Partitioning},
  author={Dao, Tri},
  booktitle={International Conference on Learning Representations (ICLR)},
  year={2024}
}

@inproceedings{pagedattn,
author = {Kwon, Woosuk and Li, Zhuohan and Zhuang, Siyuan and Sheng, Ying and Zheng, Lianmin and Yu, Cody Hao and Gonzalez, Joseph and Zhang, Hao and Stoica, Ion},
title = {Efficient Memory Management for Large Language Model Serving with PagedAttention},
year = {2023},
isbn = {9798400702297},
publisher = {Association for Computing Machinery},
address = {New York, NY, USA},
url = {https://doi.org/10.1145/3600006.3613165},
doi = {10.1145/3600006.3613165},
booktitle = {Proceedings of the 29th Symposium on Operating Systems Principles},
pages = {611–626},
numpages = {16},
location = {Koblenz, Germany},
series = {SOSP '23}
}

@inproceedings{
sun2024massive,
title={Massive Activations in Large Language Models},
author={Mingjie Sun and Xinlei Chen and J Zico Kolter and Zhuang Liu},
booktitle={First Conference on Language Modeling},
year={2024},
url={https://openreview.net/forum?id=F7aAhfitX6}
}

%%%%%%%%%%%%%%%%%%%%%%%%%%%%%%%%%%%%%%%%%%%%%%%%%%%%%%%%%%%%

\newpage
\appendix
\section*{Appendix}
% Indent entries and reduce spacing
\makeatletter
\renewcommand{\l@section}{\@dottedtocline{1}{1.5em}{2.3em}}
\makeatother
\setlength{\parskip}{4pt}

\startcontents[appendix]
\printcontents[appendix]{l}{1}[1]{%
  \setcounter{tocdepth}{2}%
  \setlength{\baselineskip}{0.8\baselineskip}%
}

\section{Variants for Capturing Value-State Magnitude and Variety}
\label{sec:app_vmag}
\begin{figure}[ht]
    \centering
    \includegraphics[width=0.8\linewidth]{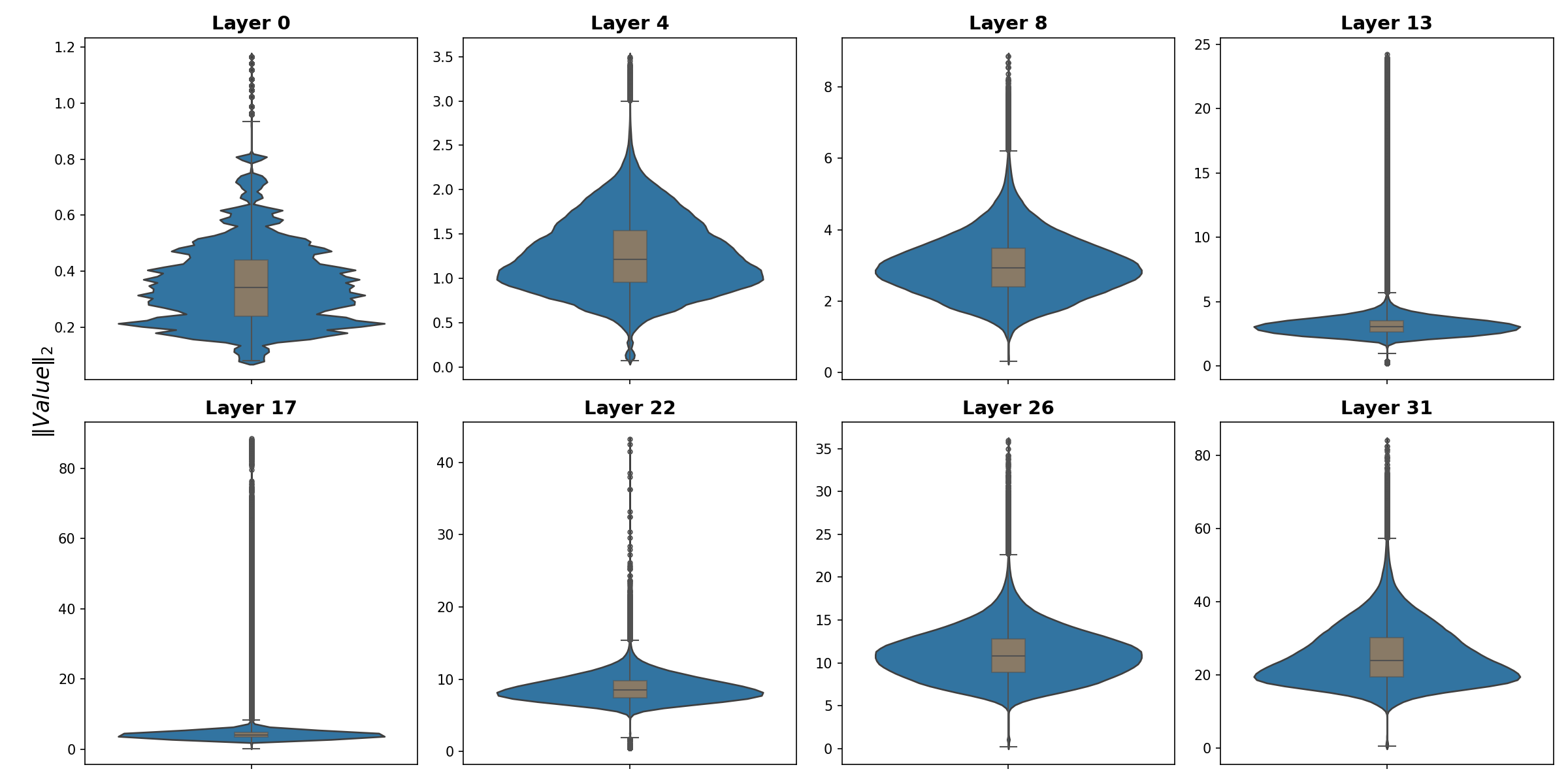}
    \caption{Layer-wise violin plots of $L_2(\vv_i) := \|\vv_i\|_2$, where the value vectors $\vv_i$ are from the full KV cache of Qwen3-4B during GSM8K generation. The distribution shows outliers that have large $L_2$ norm.}
    \label{fig:vmag_l2}
\end{figure}
\begin{figure}[ht]
    \centering
    \includegraphics[width=0.8\linewidth]{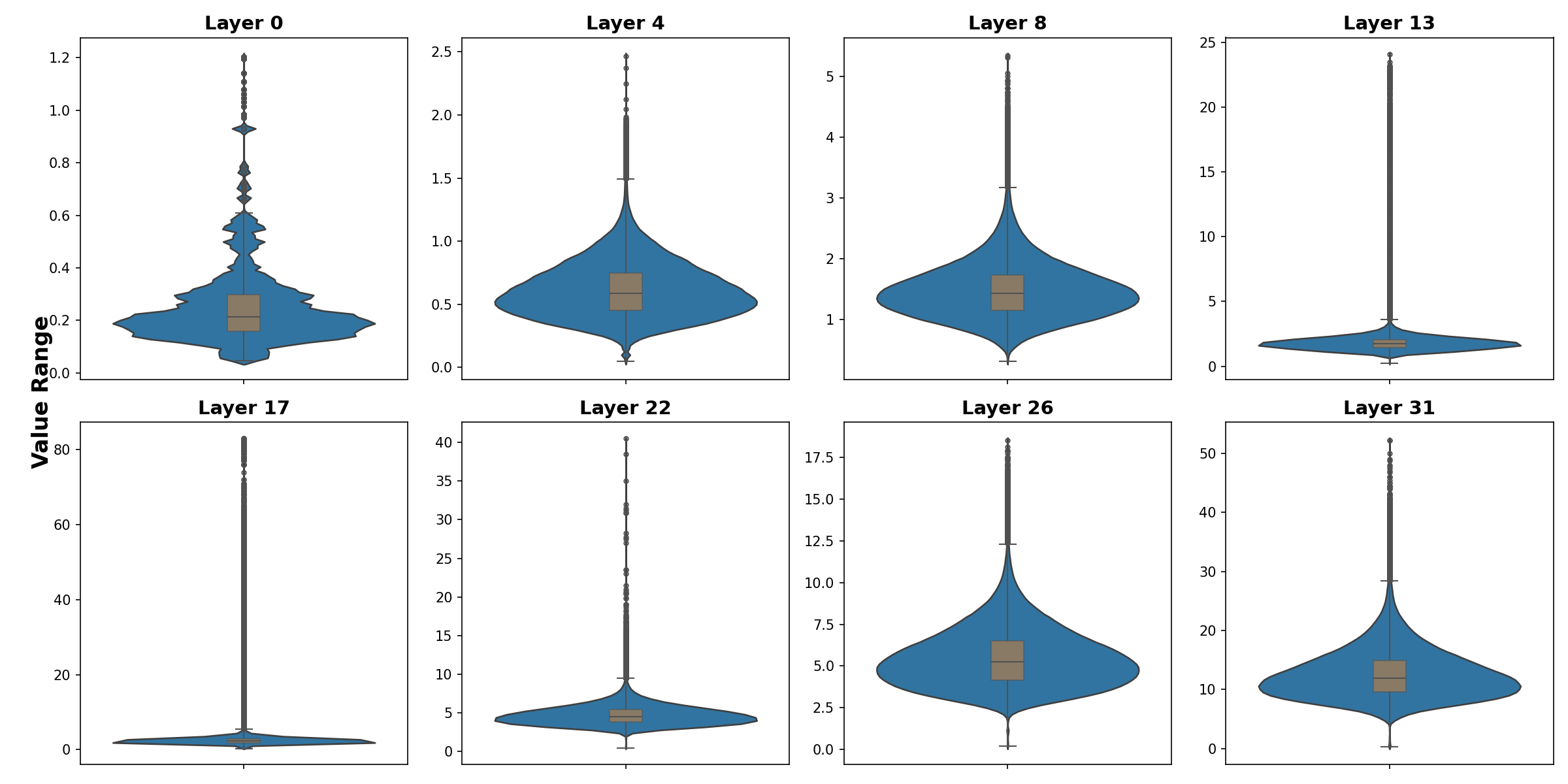}
    \caption{Layer-wise violin plots of $\mathrm{Range}(\vv_i)$, where the values $\vv_i$ are from the full KV cache of Qwen3-4B during GSM8K generation. The distribution shows outliers that have large $\mathrm{Range}$.}
    \label{fig:vmag_range}
\end{figure}
In this section, we explore different ways to compute the magnitude and variety of a value state $\vv \in \mathbb{R}^d$ in KV cache:
(1) $L_2(\vv) = \sqrt{\sum_{j=1}^{d} (v_{j})^2}$, (2) $\mathrm{Range}(\vv) \;:=\; \max_{j \in [d]} v_{j} \;-\; \min_{j \in [d]} v_{j}$, and (3) $\mathrm{Var}(\vv) = \frac{1}{d} \sum_{j=1}^{d} (\vv_{j} - \mu)^2$, where $\mu = \frac{1}{d} \sum_{j=1}^d v_j$.

First, we extract the value vectors $\vv$ from the full KV cache during the generation of Qwen3-4B on GSM8K. 
Figures~\ref{fig:vmag_l2} and \ref{fig:vmag_range} present the layer-wise distributions of $L_2(\vv)$ and $\mathrm{Range}(\vv)$, respectively.
The violin plots show prominent outliers that have large $L_2(\vv)$ and $\mathrm{Range}(\vv)$ in different layers.
We also compute the Pearson correlation between $L_2(\vv)$ and $\mathrm{Range}(\vv)$ and found that the two are highly correlated ($> 0.8$) over layers.

Next, we study whether preventing these outlier value states from eviction can maintain the accuracy under KV cache compression.
We apply the framework introduced in \Cref{sec:method} that pre-allocates dedicated value slots in the token budget.
To decide which entries to keep in the slots, we experiment with the three value-scoring variants: keeping values with the largest $L_2(\vv)$, $\mathrm{Range}(\vv)$, or $\mathrm{Var}(\vv)$, respectively.\footnote{Here, $\mathrm{Range}$ is the same as $\textsc{\ourmethod{}-AttnV}$ in the main content.}
For the remaining budget, we apply stochastic sampling on top of the attention scores in SnapKV as described in ~\Cref{sec:method}.
In \Cref{tab:cmp_vmag}, we compare the value-scoring variants on GSM8K and AIME25. 
The three variants demonstrate comparable accuracy gains over the SnapKV (attention-only) baseline.
  \begin{table*}[h]
    \centering
    \renewcommand{\arraystretch}{1.4}
    \resizebox{0.4\textwidth}{!}{
    \small
    \begin{tabular}{l c c}
    \toprule
    \textbf{Method} & \textbf{GSM8K} & \textbf{AIME25} \\
    \midrule
    SnapKV & 64.25 & 45.80 \\
    \rowcolor{ourrow} + $L_2(\vv)$ & 84.55 & 59.17 \\
    \rowcolor{ourrow} + $\mathrm{Range}(\vv)$ & 84.45 & 59.19 \\
    \rowcolor{ourrow} + $\mathrm{Var}(\vv)$ & 84.00 & 57.71 \\
    \bottomrule
    \end{tabular}
    }
    \caption{%
    Impact of value-state scoring variants on KV cache eviction accuracy. We report the results on Qwen3-4B at $4 \times$ compression. The \colorbox{ourrow}{three variants}, which capture different aspects of value-state magnitude and variety, achieve comparable performance gains over the SnapKV baseline.
    }
    \label{tab:cmp_vmag}
  \end{table*}

We choose $\Range(\vv)$ as the value scoring function for our main approach due to its connection with quantization; specifically, $\Range(\vv)$ captures the extreme values $\max \vv$ and $\min \vv$ in the vector, which is associated with the \emph{outlier} challenge in LLM quantization \citep{dettmers2022gptint}.
We show the importance of large-$\Range$ value states in \Cref{sec:v_signficance}, where evicting them causes the model to enter a nonsensical reasoning loop:

\begin{figure}[h]
    \centering
    \includegraphics[width=0.9\linewidth]{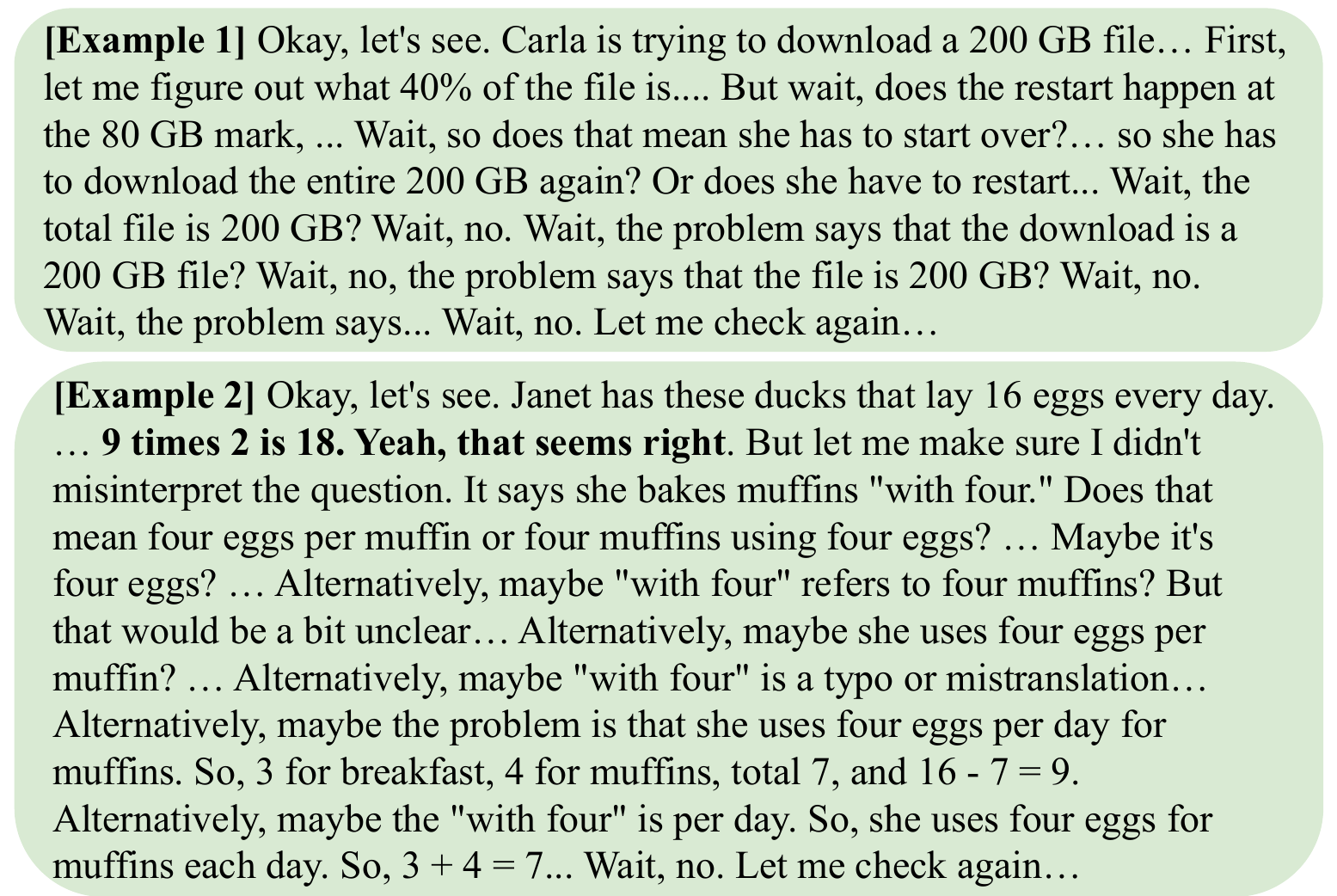}
    \caption{Examples of model outputs on GSM8K when evicting large-$\Range$ value states. In the first example, the model starts with reasonable thinking traces but falls into a repetitive loop, never reaching the correct answer. In the second example, the model generates the correct answer once (\textbf{18}), but then gets stuck in an endless loop of self-reflection and fails to reach a conclusion.}
    \label{fig:loop}
\end{figure}

\section{Token Statistics and Hyperparameters}
\label{sec:app_hyper}
\begin{table}[ht]
\begin{center}
\centering
\begin{tabular}{lcc}
\toprule
& \textbf{Prompt Tokens} & \textbf{Gen Tokens} \\
\midrule
%GSM8K & & \\
AIME25 & 198 & 17,790 \\
AIME26 & 148 & 16,642 \\
HMMT25 & 127 & 18,189 \\
MATH & 94 & 5,281 \\
LiveCodeBench-v6-medium & 557 & 11,088 \\
\bottomrule
\end{tabular}
\end{center}
\caption[Reasoning tasks token statistics.]{The average token counts of prompt (prefill) and generation (decode) of reasoning tasks. The \textbf{Gen Tokens} column represents the average number of tokens generated by the full Qwen3-4B model. The decode phase yields a significantly higher number of tokens than the prefill phase.}
\label{tab:dataset_tokens}

\end{table}
\Cref{tab:dataset_tokens} lists the average token counts of the prefill and decode phase of Qwen3-4B, respectively, showing that reasoning models may generate over $10{,}000$ tokens at decode steps for a math question that has fewer than 200 prefill tokens.
This intensifies the memory-bound constraints of the decode phase, motivating our approach to bound KV cache memory usage to a fixed cost.

Based on the statistics of \Cref{tab:dataset_tokens}, we set the token budget $K = 4096$ for AIME25, AIME26, and HMMT25, $K = 1024$ for MATH, and $K = 2048$ for GPQA-Diamond, which roughly translates to a $4\times$ KV cache compression.
For LiveCodeBench, we set $K = 2048$, which is $\sim5\times$ compression.

All the eviction methods are training-free but have hyperparameters.
We choose the best hyperparameters for each method on GSM8K and then apply them to our main experiments (\Cref{sec:main_exp}) without further tuning.
For R-KV, we experiment with different redundancy hyperparameters, $\lambda = \{0.1, 0.5, 0.9 \}$, and set $\lambda = 0.5$.
We set the Gaussian matrix rank $r=20$ for CurDKV and \ourmethoddkv{}, and the value reservation budget $N_v = K/4$ for \ourmethodattn{}.

\section{Statistical Significance}
\label{sec:app_significance}
\begin{table*}[ht]
  \centering

  \resizebox{\textwidth}{!}{
  \renewcommand{\arraystretch}{1.15}
  \small
  \begin{tabular}{l c c c c c}
  \toprule
  \textbf{Method} &
  \textbf{AIME25} & \textbf{AIME26} & \textbf{HMMT25} &
  \textbf{GPQA-D} & \textbf{MATH} \\
  \midrule
    \KVRkv           & $54.6 \pm 7.4$ & $60.2 \pm 7.4$ & $44.6 \pm 5.7$ & $59.1 \pm 3.0$ & $86.1 \pm 1.3$ \\
    \KVCur         & $53.8 \pm 7.6$ & $61.3 \pm 7.3$ & $39.6 \pm 5.3$ & $48.5 \pm 3.1$ & $75.4 \pm 1.7$ \\
   \ourmethoddkv   & $63.3 \pm 7.5$ & $68.8 \pm 6.9$ & $49.4 \pm 5.4$ & $56.8 \pm 3.0$ & $86.5 \pm 1.3$ \\
  \ourmethodattn & $64.0 \pm 7.3$ & $72.3 \pm 6.5$ & $50.0 \pm 5.6$ & $57.4 \pm 3.0$ & $85.4 \pm 1.4$ \\
  \bottomrule
  \end{tabular}
  }
  \caption{Reasoning-task accuracy (\%) with standard errors for KV cache eviction methods on Qwen3-14B under
  $\sim4\times$ cache compression.
  }
  \label{tab:se_results}
  \end{table*}
\Cref{tab:se_results} shows the average accuracy $\pm$ standard errors for KV cache eviction methods on Qwen3-14B under $\sim4\times$ cache compression. 
For each problem indexed $s \in \{1, \dots, S\}$, we sample $R$ independent generations and compute the per-problem pass@1 accuracy:
\begin{equation}
    p_s = \frac{1}{R} \sum_{r=1}^{R} \mathbf{1}[\text{sample } r \text{ of problem } s \text{ is correct}],
\end{equation}
where $R = 16$ for datasets with fewer than 100 examples and $R = 8$ otherwise. The overall pass@1 accuracy and its standard error $\mathrm{SE}$ are then:
\begin{equation}
    \bar{p} = \frac{1}{S} \sum_{s=1}^{S} p_s, \qquad \mathrm{SD} = \sqrt{\frac{1}{S-1} \sum_{s=1}^{S} (p_s - \bar{p})^2}, \qquad \mathrm{SE} = \frac{\mathrm{SD}}{\sqrt{S}} .
\end{equation}
The variability captured by the standard errors reflects both the stochasticity of the model's sampling procedure and, for \ourmethod{} methods, the randomness introduced by stochastic eviction. On tasks with larger standard errors (e.g., AIME25, AIME26), the small number of test problems (30 each) is the primary cause. 
While individual confidence intervals overlap due to small test sets, \ourmethod{} outperforms all baselines across every task–model combination. 
%Despite the overlapping confidence intervals on individual benchmarks, the improvements of \ourmethod{} over baselines are consistent in direction across all five tasks. It demonstrates that the per-task gains are not artifacts of sampling noise.

\section{Additional Results}
\label{sec:app_more_results}

Recall that in \Cref{fig:pilot}, we demonstrate that reserving $\Vslots$ slots for large-range values greatly improves GSM8K accuracy. To verify the importance of these specific value states, we conduct an ablation study where we instead reserve $\Vslots$ slots for randomly sampled KV pairs. Experimenting with $\Vslots = \{16, 32, 64\}$ yield accuracies of $65.6\%$, $63.4\%$, and $65.3\%$, respectively. These results are comparable to SnapKV ($64.3\%$) but substantially underperform the green bars in \Cref{fig:pilot}, confirming that replacing large-range values with random KV pairs is ineffective.

\Cref{tab:gpqa_budgets} shows the Qwen3-14B results of GPQA-Diamond under different token budgets.
While R-KV outperforms our \ourmethod{} methods at 2048 token budget, the accuracy gap closes at 4096 tokens ($\sim2\times$ compression).
In this setting, both R-KV and our methods nearly recover the performance of the full model.
\begin{table*}[h]
\centering
%\resizebox{\textwidth}{!}{
%\renewcommand{\arraystretch}{1.15}
\begin{tabular}{l c c | c}
\toprule
\textbf{Token Budget} & \textbf{2048} & \textbf{4096} & \textbf{Full} \\
\textbf{Compression} & $4\times $& $2\times$ & $1\times$ \\
\midrule
  \KVRkv                  & 59.09 & 63.32 & \multirow{4}{*}{65.25} \\
  \KVCur             & 48.48 & 60.54 & \\
  \ourmethoddkv          & 56.76 & 63.38 & \\
  \ourmethodattn      & 57.39 & 63.07 & \\
\bottomrule
\end{tabular}
%}
\caption{%
GPQA-D accuracy for eviction methods on Qwen3-14B under different token budgets.
}
\label{tab:gpqa_budgets}
\end{table*}

\begin{figure}[h]
    \centering
    \includegraphics[width=0.88\linewidth]{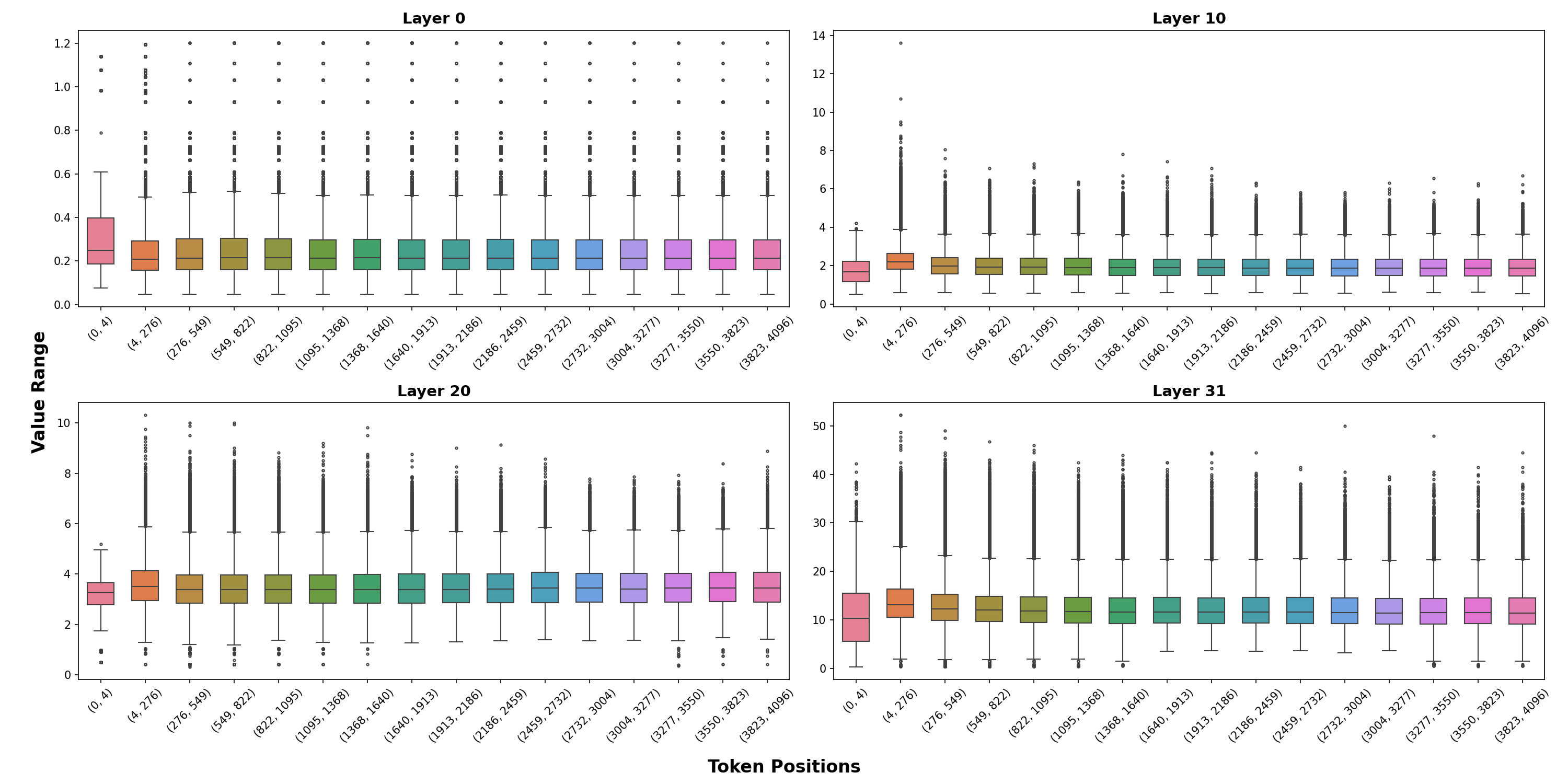}
    \caption{$\Range(\vv)$ over token position chunks, where each chunk consists of the value states of consecutive tokens. Boxplots illustrate the dynamic range (y-axis) of value states at specific layers. Token positions (x-axis) are bucketed to show how the range distribution evolves through the sequence. The first chunk contains sink tokens at position $(0,4)$. Excluding the sink tokens, the distribution of $\Range(\vv)$ does not drift as the context window grows.}
    \label{fig:chunk}
\end{figure}

In \Cref{fig:chunk}, we extract the full value cache during the generation of Qwen3-4B on GSM8K examples and plot the distributions of $\Range(\vv)$ over different chunks of consecutive tokens. 
We put the first four sink tokens \citep{su2025kvsink} into an individual chunk, which shows distinct distributions compared to the other chunks.
Except for the first layer (Layer 0), the sink tokens have a lower median $\Range(\vv)$, which corresponds to the value-state drain phenomena \citep{guo-etal-2024-attention}.
Excluding the sink tokens, the distributions of $\Range(\vv)$ are consistent as the sequence length increases.

\begin{figure}[h]
    \centering
    \includegraphics[width=1.0\linewidth]{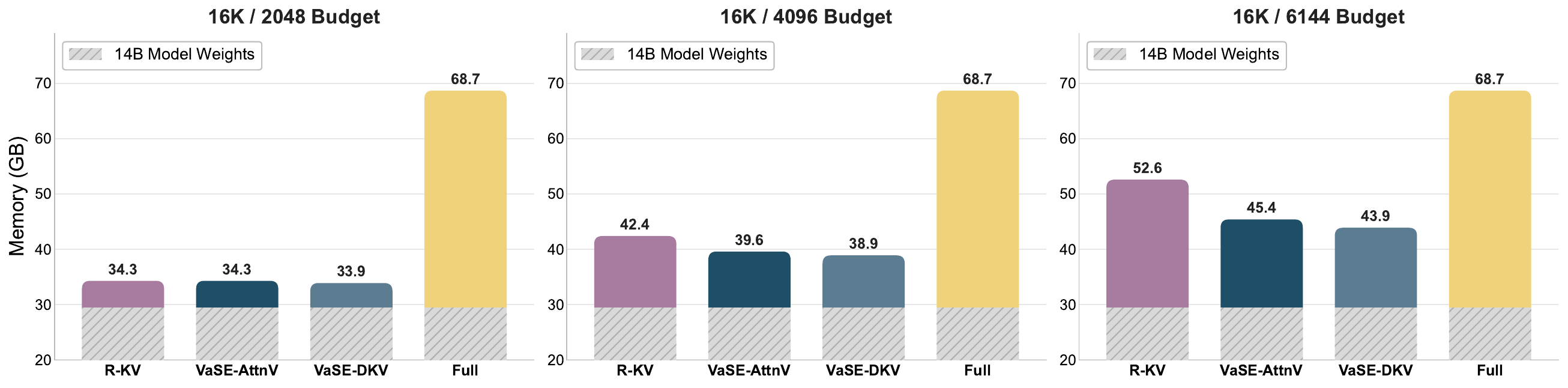}
    \caption{Peak GPU memory of Qwen3-14B at 16K output length across three KV-cache budgets $\{2048, 4096, 6144\}$. \ourmethoddkv\ is the most memory-efficient method at every budget.}
    \label{fig:memory}
\end{figure}

In \Cref{fig:memory}, we show the peak memory footprint of different methods across three KV-cache budgets $\{2048, 4096, 6144\}$ at 16K output tokens.
Overall, \ourmethoddkv $<$ \ourmethodattn $<$ \KVRkv $<$ \KVFull, showing that our \ourmethod{} methods are more memory-efficient than the \KVRkv\ baseline.
At 32K output tokens, the memory footprints of different eviction methods remain the same under the same token budgets; however, the \KVFull\ method can no longer fit a single GPU with 80GB memory.

\section{Limitations}
\label{sec:app_limitations}
Our evaluation focuses on reasoning models and the decode phase of generation, where eviction methods offer the greatest benefit. As a result, we do not evaluate on prefill-phase compression benchmarks.
However, because our method is built on fundamental observations of value-state distributions and stochastic diversity, we believe the core methodology can be extended to the prefill phase for long-prompt compression. 

We only implement our methods on Qwen3 models because our selection-based baseline SeerAttention-R only releases checkpoints for Qwen models.
Nevertheless, since our value-scoring function relies on simple $\min$ and $\max$ statistics of the value cache, our method is model-agnostic and can be easily extended to other LLM architectures.

Lastly, while our findings link value-state magnitude to quantization errors, we focus on sparse attention, and quantization method development is beyond the scope of this paper.
Future research could explore an integrated framework that combines KV cache eviction with outlier-aware quantization to push the limits of compression ratio.

\section{Compute Resources}
\label{sec:app_compute}
Our experiments were conducted on NVIDIA A100 and H100 GPUs. 
Every experiment can be run on a single GPU with 80GB memory.
Without PagedAttention and continuous batching, each task takes 10-72 GPU hours, depending on the dataset size and the average number of generated tokens.

\section{Licenses}
\label{sec:app_license}
As shown in \Cref{tab:licenses}, we include licenses for any artifacts used in this work. Copyright \copyright~MAA indicates that AIME problems are copyrighted by the Mathematical Association of America and are used here for evaluation purposes only.

% Requires in preamble:
%   \usepackage{tabularx}
%   \usepackage{xurl}
%   \usepackage{booktabs}
%   \usepackage{multirow}
%   \usepackage{hyperref}

\begin{table}[h]
\centering
\resizebox{\textwidth}{!}{%
\begin{tabularx}{1.15\textwidth}{@{}ll>{\raggedright\arraybackslash}X@{}}
\toprule
\textbf{Asset} & \textbf{License} & \textbf{Source} \\
\midrule
\multicolumn{3}{@{}l}{\textit{Models}} \\[2pt]
Qwen3-4B                & Apache 2.0              & \href{https://huggingface.co/Qwen/Qwen3-4B}{\nolinkurl{huggingface.co/Qwen/Qwen3-4B}} \\
Qwen3-14B               & Apache 2.0              & \href{https://huggingface.co/Qwen/Qwen3-14B}{\nolinkurl{huggingface.co/Qwen/Qwen3-14B}} \\
\midrule
\multicolumn{3}{@{}l}{\textit{Datasets \& Benchmarks}} \\[2pt]
GSM8K                    & MIT                     & \href{https://github.com/openai/grade-school-math}{\nolinkurl{github.com/openai/grade-school-math}} \\
MATH                     & MIT                     & \href{https://github.com/hendrycks/math}{\nolinkurl{github.com/hendrycks/math}} \\
GPQA-Diamond             & CC-BY-4.0               & \href{https://huggingface.co/datasets/Idavidrein/gpqa}{\nolinkurl{huggingface.co/datasets/Idavidrein/gpqa}} \\
LiveCodeBench-v6         & MIT (code) / CC (data)  & \href{https://github.com/LiveCodeBench/LiveCodeBench}{\nolinkurl{github.com/LiveCodeBench/LiveCodeBench}} \\
AIME 2025 / 2026         & Copyright \copyright~MAA       & \href{https://artofproblemsolving.com/wiki}{\nolinkurl{artofproblemsolving.com/wiki}} \\
\multirow[t]{2}{*}{HMMT 2025} & \multirow[t]{2}{*}{CC-BY-NC-SA 4.0} & \href{https://huggingface.co/datasets/MathArena/hmmt_feb_2025}{\nolinkurl{huggingface.co/datasets/MathArena/hmmt_feb_2025}} \\
                         &                         & \href{https://huggingface.co/datasets/MathArena/hmmt_nov_2025}{\nolinkurl{huggingface.co/datasets/MathArena/hmmt_nov_2025}} \\
\midrule
\multicolumn{3}{@{}l}{\textit{Code \& Baselines}} \\[2pt]
SnapKV                   & CC-BY-4.0               & \href{https://openreview.net/forum?id=poE54GOq2l}{\nolinkurl{openreview.net/forum?id=poE54GOq2l}} \\
R-KV                     & CC-BY-4.0               & \href{https://openreview.net/forum?id=2jwAjomEDB}{\nolinkurl{openreview.net/forum?id=2jwAjomEDB}} \\
CurDKV                   & CC-BY-4.0               & \href{https://openreview.net/forum?id=klmc4fwPLd}{\nolinkurl{openreview.net/forum?id=klmc4fwPLd}} \\
SeerAttention-R          & CC-BY-4.0               & \href{https://openreview.net/forum?id=c5BOcHM6J8}{\nolinkurl{openreview.net/forum?id=c5BOcHM6J8}} \\
HQQ                      & Apache 2.0              & \href{https://github.com/dropbox/hqq}{\nolinkurl{github.com/dropbox/hqq}} \\
\bottomrule
\end{tabularx}%
}
\vspace{1ex}
\caption{Assets and their licenses.}
\label{tab:licenses}
\end{table}

\section{Broader Impacts}
\label{sec:app_broader}
This work proposes a training-free KV cache eviction method that reduces the memory footprint of reasoning models during inference. The primary positive impact is enabling more efficient deployment of large language models, reducing hardware requirements and energy consumption. This can broaden access to reasoning-capable models for resource-constrained practitioners and organizations.

As a general-purpose inference optimization, \ourmethod{} does not introduce new model capabilities or alter model outputs beyond the approximation inherent in cache compression. It therefore does not directly raise new ethical concerns beyond those already associated with the underlying language models. We do not foresee specific negative societal impacts arising from this work.

\makeatletter\if@preprint\else
\newpage
\section*{NeurIPS Paper Checklist}
\addcontentsline{toc}{section}{\protect\numberline{$\star$}NeurIPS Paper Checklist}
%%% BEGIN INSTRUCTIONS %%%
The checklist is designed to encourage best practices for responsible machine learning research, addressing issues of reproducibility, transparency, research ethics, and societal impact. Do not remove the checklist: {\bf The papers not including the checklist will be desk rejected.} The checklist should follow the references and follow the (optional) supplemental material.  The checklist does NOT count towards the page
limit. 

Please read the checklist guidelines carefully for information on how to answer these questions. For each question in the checklist:
\begin{itemize}
    \item You should answer \answerYes{}, \answerNo{}, or \answerNA{}.
    \item \answerNA{} means either that the question is Not Applicable for that particular paper or the relevant information is Not Available.
    \item Please provide a short (1--2 sentence) justification right after your answer (even for \answerNA). 
   % \item {\bf The papers not including the checklist will be desk rejected.}
\end{itemize}

{\bf The checklist answers are an integral part of your paper submission.} They are visible to the reviewers, area chairs, senior area chairs, and ethics reviewers. You will also be asked to include it (after eventual revisions) with the final version of your paper, and its final version will be published with the paper.

The reviewers of your paper will be asked to use the checklist as one of the factors in their evaluation. While \answerYes{} is generally preferable to \answerNo{}, it is perfectly acceptable to answer \answerNo{} provided a proper justification is given (e.g., error bars are not reported because it would be too computationally expensive'' or ``we were unable to find the license for the dataset we used''). In general, answering \answerNo{} or \answerNA{} is not grounds for rejection. While the questions are phrased in a binary way, we acknowledge that the true answer is often more nuanced, so please just use your best judgment and write a justification to elaborate. All supporting evidence can appear either in the main paper or the supplemental material, provided in appendix. If you answer \answerYes{} to a question, in the justification please point to the section(s) where related material for the question can be found.

IMPORTANT, please:
\begin{itemize}
    \item {\bf Delete this instruction block, but keep the section heading ``NeurIPS Paper Checklist"},
    \item  {\bf Keep the checklist subsection headings, questions/answers and guidelines below.}
    \item {\bf Do not modify the questions and only use the provided macros for your answers}.
\end{itemize}

%%% END INSTRUCTIONS %%%

\begin{enumerate}

\item {\bf Claims}
    \item[] Question: Do the main claims made in the abstract and introduction accurately reflect the paper's contributions and scope?
    \item[] Answer: \answerYes{} % Replace by \answerYes{}, \answerNo{}, or \answerNA{}.
    \item[] Justification: Our claims are backed by our experimental results in \S~\ref{sec:experiments}.

\item {\bf Limitations}
    \item[] Question: Does the paper discuss the limitations of the work performed by the authors?
    \item[] Answer: \answerYes{} % Replace by \answerYes{}, \answerNo{}, or \answerNA{}.
    \item[] Justification: We talk about limitations in \S~\ref{sec:app_limitations}.

\item {\bf Theory assumptions and proofs}
    \item[] Question: For each theoretical result, does the paper provide the full set of assumptions and a complete (and correct) proof?
    \item[] Answer: \answerNA{} % Replace by \answerYes{}, \answerNo{}, or \answerNA{}.
    \item[] Justification: \answerNA{}

    \item {\bf Experimental result reproducibility}
    \item[] Question: Does the paper fully disclose all the information needed to reproduce the main experimental results of the paper to the extent that it affects the main claims and/or conclusions of the paper (regardless of whether the code and data are provided or not)?
    \item[] Answer: \answerYes{} % Replace by \answerYes{}, \answerNo{}, or \answerNA{}.
    \item[] Justification: We disclose all of our experiment details in \S~\ref{sec:experiments} and in Appendix \S~\ref{sec:app_hyper}.

\item {\bf Open access to data and code}
    \item[] Question: Does the paper provide open access to the data and code, with sufficient instructions to faithfully reproduce the main experimental results, as described in supplemental material?
    \item[] Answer: \answerYes{} % Replace by \answerYes{}, \answerNo{}, or \answerNA{}.
    \item[] Justification: We will release codes upon acceptance.

\item {\bf Experimental setting/details}
    \item[] Question: Does the paper specify all the training and test details (e.g., data splits, hyperparameters, how they were chosen, type of optimizer) necessary to understand the results?
    \item[] Answer: \answerYes{} % Replace by \answerYes{}, \answerNo{}, or \answerNA{}.
    \item[] Justification: All experiment details are disclosed in \S~\ref{sec:app_hyper}.

\item {\bf Experiment statistical significance}
    \item[] Question: Does the paper report error bars suitably and correctly defined or other appropriate information about the statistical significance of the experiments?
    \item[] Answer: \answerYes{} % Replace by \answerYes{}, \answerNo{}, or \answerNA{}.
    \item[] Justification: We provide standard errors in \Cref{tab:se_results}.

\item {\bf Experiments compute resources}
    \item[] Question: For each experiment, does the paper provide sufficient information on the computer resources (type of compute workers, memory, time of execution) needed to reproduce the experiments?
    \item[] Answer: \answerYes{} % Replace by \answerYes{}, \answerNo{}, or \answerNA{}.
    \item[] Justification: We talk about compute details in \S~\ref{sec:app_compute}.
    
\item {\bf Code of ethics}
    \item[] Question: Does the research conducted in the paper conform, in every respect, with the NeurIPS Code of Ethics \url{https://neurips.cc/public/EthicsGuidelines}?
    \item[] Answer: \answerYes{} % Replace by \answerYes{}, \answerNo{}, or \answerNA{}.
    \item[] Justification: The authors have read the NeurIPS Code of Ethics and made sure the paper follows the NeurIPS Code of Ethics in every aspect.

\item {\bf Broader impacts}
    \item[] Question: Does the paper discuss both potential positive societal impacts and negative societal impacts of the work performed?
    \item[] Answer: \answerYes{} % Replace by \answerYes{}, \answerNo{}, or \answerNA{}.
    \item[] Justification: We discuss broader impacts in \S~\ref{sec:app_broader}.
    
\item {\bf Safeguards}
    \item[] Question: Does the paper describe safeguards that have been put in place for responsible release of data or models that have a high risk for misuse (e.g., pre-trained language models, image generators, or scraped datasets)?
    \item[] Answer: \answerNA{} % Replace by \answerYes{}, \answerNo{}, or \answerNA{}.
    \item[] Justification: This paper is about KV cache eviction which works on existing open-weight models, where we believe the safeguards have been done by the model providers.

\item {\bf Licenses for existing assets}
    \item[] Question: Are the creators or original owners of assets (e.g., code, data, models), used in the paper, properly credited and are the license and terms of use explicitly mentioned and properly respected?
    \item[] Answer: \answerYes{} % Replace by \answerYes{}, \answerNo{}, or \answerNA{}.
    \item[] Justification: We talk about licenses and datasets in \S~\ref{sec:app_license}.

\item {\bf New assets}
    \item[] Question: Are new assets introduced in the paper well documented and is the documentation provided alongside the assets?
    \item[] Answer: \answerNA{} % Replace by \answerYes{}, \answerNo{}, or \answerNA{}.
    \item[] Justification: \answerNA{}

\item {\bf Crowdsourcing and research with human subjects}
    \item[] Question: For crowdsourcing experiments and research with human subjects, does the paper include the full text of instructions given to participants and screenshots, if applicable, as well as details about compensation (if any)? 
    \item[] Answer: \answerNA{} % Replace by \answerYes{}, \answerNo{}, or \answerNA{}.
    \item[] Justification: \answerNA{}

\item {\bf Institutional review board (IRB) approvals or equivalent for research with human subjects}
    \item[] Question: Does the paper describe potential risks incurred by study participants, whether such risks were disclosed to the subjects, and whether Institutional Review Board (IRB) approvals (or an equivalent approval/review based on the requirements of your country or institution) were obtained?
    \item[] Answer: \answerNA{} % Replace by \answerYes{}, \answerNo{}, or \answerNA{}.
    \item[] Justification: \answerNA{}

\item {\bf Declaration of LLM usage}
    \item[] Question: Does the paper describe the usage of LLMs if it is an important, original, or non-standard component of the core methods in this research? Note that if the LLM is used only for writing, editing, or formatting purposes and does \emph{not} impact the core methodology, scientific rigor, or originality of the research, declaration is not required.
    %this research? 
    \item[] Answer: \answerNA{} % Replace by \answerYes{}, \answerNo{}, or \answerNA{}.
    \item[] Justification: LLMs are used only for writing, editing, or formatting purposes. 

\end{enumerate}
\fi\makeatother

\end{document}